\documentclass{article}
\pdfoutput=1
\usepackage{arxiv,times}
\usepackage{times}
\usepackage{url}
\usepackage{graphicx, subfigure, multirow}
\usepackage{amsmath, amsfonts}
\usepackage{latexsym}
\usepackage{enumitem}
\usepackage{lipsum}
\usepackage[linesnumbered,ruled]{algorithm2e}

\title{Geometry of Polysemy}

\author{Jiaqi Mu, Suma Bhat, Pramod Viswanath \\
Department of Electrical and Computer Engineering\\
University of Illinois at Urbana Champaign\\
Urbana, IL 61801, USA \\
\texttt{\{jiaqimu2,spbhat2,pramodv\}@illinois.edu}}

\begin{document}

\maketitle

\begin{abstract}
  Vector representations of words have heralded a transformational approach to classical problems in NLP; the most popular example is word2vec. However, a single vector does not suffice to model the polysemous nature of many (frequent) words, i.e., words with multiple meanings.    In this paper, we propose a three-fold approach for {\em unsupervised} polysemy modeling: (a) context representations, (b) sense induction and disambiguation and (c) lexeme (as a word and sense pair) representations. A key feature of our work is the finding that  a sentence containing a target word is  well represented by a low rank {\em subspace},  instead of a point in a vector  space. We then show that the subspaces associated with a particular sense of the target word tend to {\em intersect} over a line (one-dimensional subspace), which we use to disambiguate senses using a  clustering algorithm that harnesses the Grassmannian geometry of the representations. The disambiguation algorithm, which we call $K$-Grassmeans, leads to a procedure to label the different senses of the target word in the corpus -- yielding lexeme vector representations, all in an unsupervised manner starting from a large (Wikipedia) corpus in English. Apart from several prototypical  target (word,sense) examples and a host of empirical studies to intuit and justify  the various geometric representations,  we validate our algorithms on standard sense induction and disambiguation datasets and present new state-of-the-art results.
\end{abstract}

\section{Introduction}
Distributed representations are embeddings of words in a real vector space, achieved  via an appropriate function that models the interaction between neighboring words in sentences  (e.g.: neural networks \cite{DBLP:journals/jmlr/BengioDVJ03,DBLP:conf/interspeech/MikolovKBCK10,DBLP:conf/acl/HuangSMN12}, log-bilinear models \cite{DBLP:conf/icml/MnihH07,DBLP:journals/corr/abs-1301-3781}, co-occurrence statistics \cite{DBLP:conf/emnlp/PenningtonSM14,DBLP:conf/nips/LevyG14}). Such an approach has been strikingly successful in capturing the syntactic and semantic similarity between words (and pairs of words), via simple linear algebraic relations between their corresponding vector representations.  On the other hand, the {\em polysemous}  nature of words, i.e., the phenomenon of the same surface form representing multiple senses, is a central feature of the creative process embodying all natural languages. For example, a large, tall machine used for moving heavy objects and a tall, long-legged, long-necked bird both share the same surface form ``crane''. A vast majority of words, especially frequent ones, are polysemous, with each word taking on anywhere from two to a dozen different senses in many natural languages. For instance, WordNet collects 26,896 polysemous English words with an average of 4.77 senses each \cite{DBLP:journals/cacm/Miller95}. Naturally, a single vector embedding does not appropriately represent a polysemous word.   

There are currently two approaches to address the polysemy issue:
\begin{itemize}
  \item Sense specific representation learning \cite{DBLP:conf/emnlp/ChenLS14,DBLP:conf/acl/RotheS15}, usually aided by hand-crafted lexical resources such as WordNet \cite{DBLP:journals/cacm/Miller95};
  \item Unsupervised sense induction and sense/lexeme representation learning by inferring the senses directly from text \cite{DBLP:conf/acl/HuangSMN12,DBLP:conf/emnlp/NeelakantanSPM14,DBLP:conf/emnlp/LiJ15,DBLP:journals/corr/AroraLLMR16}.
\end{itemize}
Since hand-crafted lexical resources sometimes do not reflect the actual meaning of a target word in a given context \cite{DBLP:journals/csl/Veronis04} and, more importantly, such resources are lacking in many languages (and their creation draws upon intensive expert human resources), we focus on the second approach in  this paper; such an approach is inherently scalable and potentially plausible with the right set of ideas. Indeed, a human expects the contexts to cue in on the particular sense of a specific word, and  successful unsupervised sense representation and sense extraction algorithms would represent progress in the broader area of representation of natural language. Such are the goals of this work.   

 Firth1957's hypothesis -- a word is characterized by the company it keeps \cite{Firth1957} -- has motivated the development  of single embeddings for  words,  but also suggests that multiple senses for a target word could be inferred from its contexts (neighboring words within the sentence). This task is naturally broken into three related questions:  (a) how to {\em represent} contexts (neighboring words of the target word); (b) how to {\em induce} word senses (partition  instances of contexts into groups where the target word is used in the same sense within each group) and (c) how to represent lexemes (word and sense pairs) by   vectors.

Existing works address these questions by exploring the latent structure of contexts. In an inspired work, \cite{DBLP:journals/corr/AroraLLMR16} hypothesizes that the global word representation is a linear combination of its sense representations, models the contexts by a finite number of discourse atoms, and recovers the sense representations via sparse coding of all the vectors of the vocabulary (a global fit). Other works perform a local context-specific sense induction: \cite{DBLP:conf/emnlp/LiJ15} introduces a sense-based language model to disambiguate word senses and to learn lexeme representations by incorporating the Chinese restaurant process, \cite{DBLP:conf/naacl/ReisingerM10} and \cite{DBLP:conf/acl/HuangSMN12}  label the word senses by clustering the contexts based on the average of the context word embeddings and learn lexeme representations using the labeled corpus. \cite{DBLP:conf/emnlp/NeelakantanSPM14} retains the representation of contexts by the average of the word vectors, but improves the previous approach by jointly learning the lexeme vectors and the cluster centroids.

{\bf Grassmannian Model}: We depart from the linear latent models in these prior works by presenting a {\em nonlinear} (Grassmannian) geometric property of contexts.  We empirically observe and hypothesize that the 
 context word representations surrounding a target word reside roughly in a low dimensional {\em subspace}. Under this hypothesis, a specific sense representation for a target word should reside in all the subspaces of the contexts where the word means this sense. Note that  these subspaces need not cluster at all: a word such as ``launch" in the sense of ``beginning or initiating a new endeavor" could be used in a large variety of contexts. Nevertheless, our hypothesis that large semantic units (such as sentences) reside in low dimensional subspaces implies that  the subspaces of all contexts where the target word shares the same meaning should {\em intersect} non-trivially. This further implies that there exists a direction (one dimensional subspace) that is very close to all subspaces and we treat such an intersection vector as the representation of a group of subspaces. 

Following this intuition, we propose a three-fold approach to deal with the three central questions posed above. 
\begin{itemize}
  \item {\bf Context Representation}: we define the context for a target word to be a set of left $W$ and right $W$ non-functional words of the target word ($W\approx 10$ in our experiments), including the target word itself, and represent it by a low-dimensional subspace spanned by its context word representations; 
  \item {\bf Sense Induction and Disambiguation}: we induce word senses from their contexts by partitioning multiple  context instances into groups, where the target word has the same sense within each group. Each group is associated with a representation -- the intersection direction of the group -- found via $K$-Grassmeans, a novel clustering method that harnesses the geometry of subspaces. Finally, we  disambiguate word senses for new context instances using the respecive group representations;
  \item {\bf Lexeme Representation}: the lexeme representations can be obtained by running an off-the-shelf word embedding algorithm on a labeled corpus. We label the corpus through hard decisions (involving {\em erasure} labels) and soft decisions (probabilistic labels), motivated by analogous successful approaches to decoding of turbo and LDPC codes in wireless communication. 
\end{itemize}

{\bf Experiments}: The lexical aspect of our algorithm (i.e., senses can be induced and disambiguated {\em individually} for each word) as well as the novel geometry  (subspace intersection vectors) jointly allow us to capture subtle shades of senses. For instance, in ``Can you hear me? You're on the {\bf air}.  One of the great moments of live television, isn't it?'', our representation is able to capture the occurrence of ``air'' to mean ``live event on camera''. In contrast, with a global fit such as that  in \cite{DBLP:journals/corr/AroraLLMR16} the senses are inherently limited in the number and type of ``discourse atoms'' that can be captured.   

As a quantitative demonstration of the latent geometry captured by our methods,  we evaluate the proposed induction algorithm on standard Word Sense Induction (WSI) tasks. Our algorithm outperforms state-of-the-art on two datasets: (a) SemEval-2010 Task 14 \cite{DBLP:conf/semeval/ManandharKDP10} whose word senses are obtained from OntoNotes \cite{DBLP:conf/naacl/PradhanX09}; and (b) a custom-built dataset built by repurposing the  polysemous dataset of \cite{DBLP:journals/corr/AroraLLMR16}. 
In terms of lexeme vector embeddings, our  representations have evaluations comparable to state-of-the-art on  standard tasks -- the word similarity task of SCWS \cite{DBLP:conf/acl/HuangSMN12} -- and significantly better on a subset of the SCWS dataset which focuses on polysemous target words and the ``police lineup" task of  \cite{DBLP:journals/corr/AroraLLMR16}.

We summarize our contributions below:
\begin{itemize}
  \item We observe a new geometric property of context words and use it to represent contexts by the low-dimensional subspaces spanned by their word vector representations;
  \item We use the geometry of the subspaces in conjunction with unsupervised clustering methods to propose a sense induction and disambiguation algorithm;
  \item We introduce a new dataset for the WSI task which includes 50 polysemous words and 6,567 contexts. The word senses in this dataset are coarser and more human-interpretable than those in previous WSI datasets. 
\end{itemize}

\section{Context Representation}
\label{sec:context-representation}

Contexts refer to  entire sentences or (long enough) consecutive blocks of words in sentences surrounding a target word. Efficient distributed vector representations for sentences and paragraphs are active topics of research in the literature (\cite{DBLP:conf/icml/LeM14,DBLP:conf/acl/TaiSM15,DBLP:conf/emnlp/Kim14}), with much emphasis on appropriately relating the individual word embeddings with those of the sentences (and paragraphs) they reside in.  The scenario of contexts studied here is similar in the sense that they constitute long semantic units similar to sentences, but  different in that we are considering semantic units that all have a {\em common} target word residing inside them.  Instead of a straightforward application of existing literature on sentence (and paragraph) vector embeddings to our setting, we deviate and propose a {\em non-vector} space representation;  such a representation is central to the results of this paper and is best motivated by the following simple experiment. 
 
\begin{figure}[h]
\centering
{\includegraphics[width=0.4\textwidth]{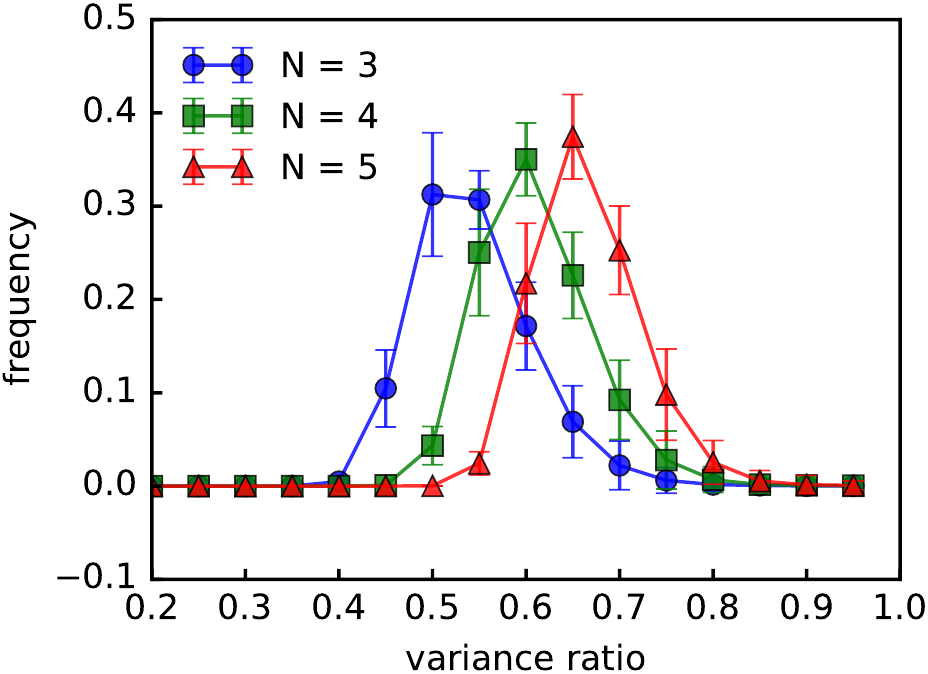}}
\caption{An experiment to study the linear algebraic structure of word senses.\label{fig:lowrank}}
\end{figure}

Given a random word and a set of its contexts (culled from the set of all sentences where the target word appears), we use principle component analysis (PCA) to project the context word embeddings for every context into an $N$-dimensional subspace and measure the low dimensional nature of context word embeddings. We randomly sampled 500 words whose occurrence (frequency) is larger than 10,000, extracted their contexts from Wikipedia, and plotted the histogram of the variance ratios being captured by rank-$N$ PCA in Figure \ref{fig:lowrank}(a) for  $N=3,4,5$. We make the following observations: 
even rank-3 PCA captures at least 45\% of the energy (i.e., variance ratio) of the context word representations and rank-$4$ PCA can capture at least half of the energy almost surely. As comparison, we note that the average the number of context words is roughly 21 and a rank-4 PCA over a random collection of 21 words would be expected to capture only 20\% of the energy (this calculation is justified because word vectors have been observed to possess a spatial isotropy property \cite{TACL742}). All word vectors were trained on the Wikipedia corpus with dimension $d=300$ using the  skip-gram program of word2vec \cite{DBLP:journals/corr/abs-1301-3781}.

This experiment immediately suggests the low-dimensional nature of contexts, and that the contexts be represented in the space of subspaces, i.e., the Grassmannian manifold:  we represent a context $c$ (as a multiset of words) by a point in the Grassmannian manifold -- a subspace (denoted by $S(c)$) spanned by its top $N$ principle components (denoted by $\{u_n(c)\}_{n=1}^N$), i.e.,
\begin{align*}
	S(c) = \left\{\sum_{n=1}^N \alpha_n u_n(c): \alpha_n \in \mathbb{R}\right\}.
\end{align*}
A detailed algorithm chart for context representations is provided in  Appendix~\ref{app:algo1}, for completeness. 

\section{Sense Induction and Disambiguation}

We now turn to sense induction, a basic task that explores polysemy: in this task, a set of sentences (each containing a common target word)  have to be partitioned such that the target word is used in the same sense in all the sentences within each partition. The number of partitions relates to the number of senses being identified for the target word. The geometry of the subset representations plays a key role in our algorithm and we start with this next.  

\subsection{Geometry of Polysemy}
Consider a target word $w$ and a context sentence $c$ containing this word $w$. The empirical experiment from the previous section allows us to represent $c$ by a $N$-dimensional subspace of the vectors of the words in $c$. Since $N$ (3$\sim$5 in our experiments) is much smaller than the number of words in $c$ (21, on average), one suspects that the representation associated with $c$ wouldn't change very much if the target word $w$ were expurgated from it, i.e., 
$$S(c) \approx S(c \setminus w).$$
On the other hand $v(w)$ (the vector representation of the target word $w$) perhaps has a fairly large intersection with $S(c)$ and thus also with $S(c\setminus w)$. Putting these two observations together, one arrives at the following hypothesis, in the context of {\em monosemous} target words: 
\begin{quote}
{\bf  Intersection Hypothesis}: the target word vector $v(w)$ should reside in the {\em intersection} of $S(c \setminus w)$, where the intersection is over all its contexts $c$. 
\end{quote}
The reason why this hypothesis is made in the context of monosemous words is that in this case the word vector representation is ``pure'', while polysemous words are really different words with the same (lexical) surface form.

\begin{figure}[h]
\centering
\subfigure[distance between word representation and its subspaces]
{\includegraphics[width=0.4\textwidth]{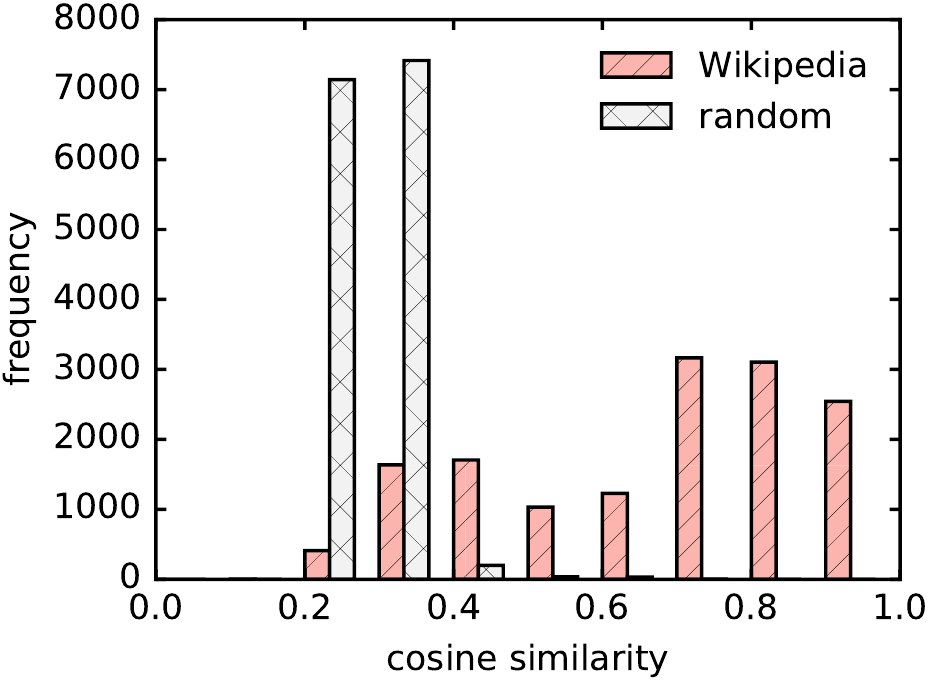}}
\subfigure[contexts of ``typhoon'']
{
\includegraphics[width=0.4\textwidth]{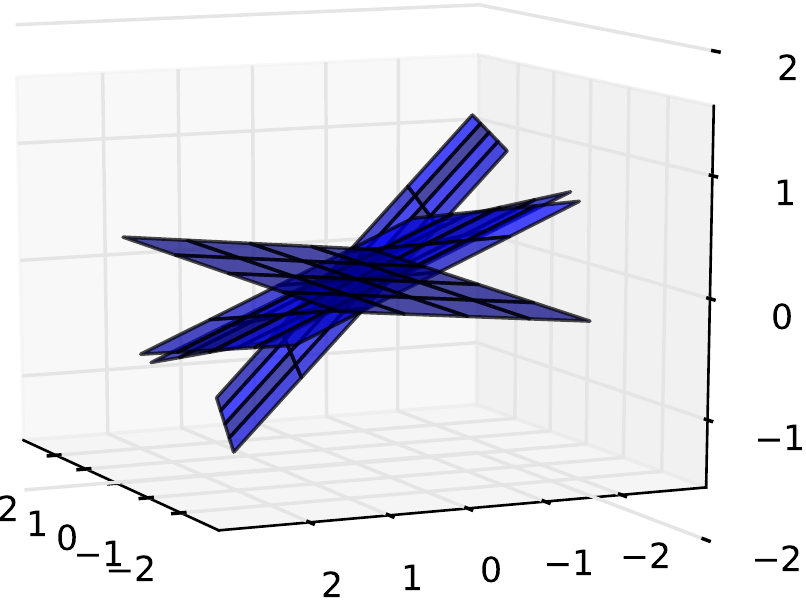}
}
\caption{The geometry of contexts for monosemy.\label{fig:monosemy}}
\end{figure}

{\bf Empirical Validation of Intersection Hypothesis}: We illustrate the intersection phenomenon via another experiment. Consider a monosemous word ``typhoon'' and consider all contexts in Wikipedia corpus where this word appears (there are 14,829  contexts and a few sample contexts are provided in Table \ref{tb:monosemy}). We represent each context by the  rank-$N$ PCA subspace of all the vectors (with $N=3$) associated with the words  in the context and consider their  intersection. Each of these subspaces is $3\times d$ dimensional (where $d = 300$ is the dimension of the word vectors).  We  find that cosine similarity (normalized projection distance) between the vector associated with ``typhoon" and each context subspace is very small: the average is 0.693 with standard deviation 0.211.  For comparison,  we randomly sample 14,829 contexts and find the average is 0.305 with standard deviation 0.041 (a detailed histogram is provided in Figure~\ref{fig:monosemy}(a)). This  corroborates with the hypothesis  that the target word vector is in the intersecton of the context subspaces. A visual representation of this geometric phenomenon is in Figure \ref{fig:monosemy}(b), where we have projected the $d$-dimensional word representations into  3-dimensional vectors and use these 3-dimensional word vectors to get the subspaces for contexts (we set $N=2$ here for visualization) in Table~\ref{tb:monosemy}, and plot the subspaces as $2$-dimensional planes. From Figure \ref{fig:monosemy},  we can see that all the context subspaces roughly intersect at a common direction, thus empirically justifying the intersection hypothesis.

\begin{table}[htbp]
  \begin{center}
    \begin{tabular}{|p{0.92\textwidth}|}
      \hline
      \multicolumn{1}{|c|}{\bf ``typhoon''}  \\
      \hline
      powerful typhoon that affected southern japan in july it was the sixth named storm and second typhoon of the pacific {\bf typhoon} season originating from an area of low pressure near wake island on july the precursor to maon gradually developed
 \\
      \hline
     typhoon ineng was a powerful typhoon that affected southern japan in july it was the sixth named storm and second {\bf typhoon} of the pacific typhoon season originating from an area of low pressure near wake island on july the precursor
 \\
      \hline
      crossing through a gulf of the south china sea patsy weakened to a mph kmh tropical storm before the joint {\bf typhoon} warning center ceased following the system on the morning of august as it made landfall near the city of
 \\
      \hline
      bay were completely wiped out while all airplanes at the local airports were damaged this is the first active pacific {\bf typhoon} season on record typhoon barbara formed on march and moved west it strengthened briefly to a category three with
\\
      \hline
    \end{tabular}
  \end{center}
  \caption{Contexts containing a monosemous  word ``typhoon''\label{tb:monosemy}.}
\end{table}

{\bf Recovering the Intersection Direction}: An algorithmic approach to robustly discover the intersection direction involves  finding that direction vector that is ``closest" to all subspaces; we propose doing so   by solving the following optimization problem: 
\begin{align}
  \hat{u}(w) = \arg \min_{\|u\|=1} \sum_{w \in c} d(u, S(c \setminus w))^2, \label{eq:mono-opt}
\end{align}
where $d(v, S)$ is the shortest $\ell_2$-distance between $u$ and subspace $S$, i.e.,
\begin{align*}
  d(u, S) = \sqrt{\|u\|^2 - \sum_{n=1}^N \left(u^{\rm T}u_n\right)^2},
\end{align*}
where $u_1, \ldots, u_N$ are $N$ orthonormal basis vectors for subspace $S$. Thus \eqref{eq:mono-opt} is equivalent to,
\begin{align}
  \hat{u}(w) = \arg \max_{\|u\|=1} \sum_{w\in c} \sum_{n=1}^N  \left(u^{\rm T}u_n(c \setminus w)\right)^2,
\end{align}
which can be solved by taking the first principle component of $\{u_n(c \setminus w)\}_{w\in c, n=1,\ldots,N}$.

\begin{figure*}[h]
  \centering
  \subfigure[crane: machine]
  {
  \includegraphics[width=0.3\textwidth]{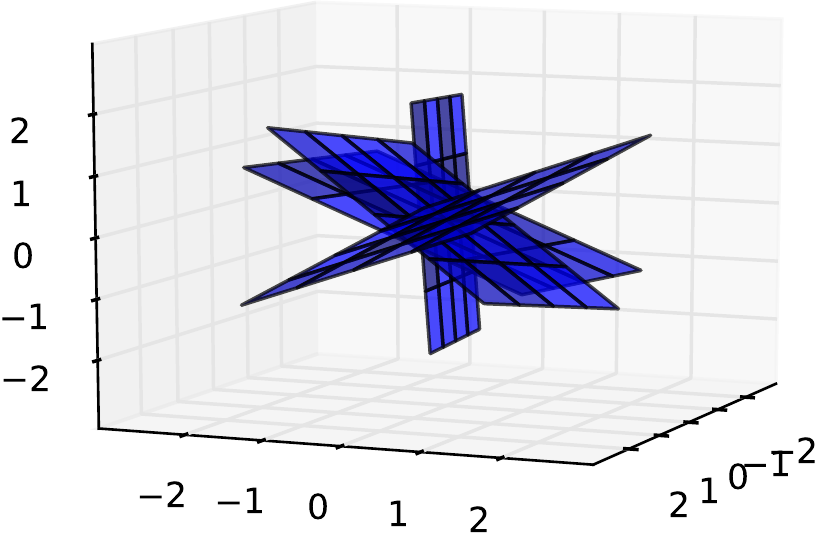}
  }
  \subfigure[crane: bird]
  {
  \includegraphics[width=0.3\textwidth]{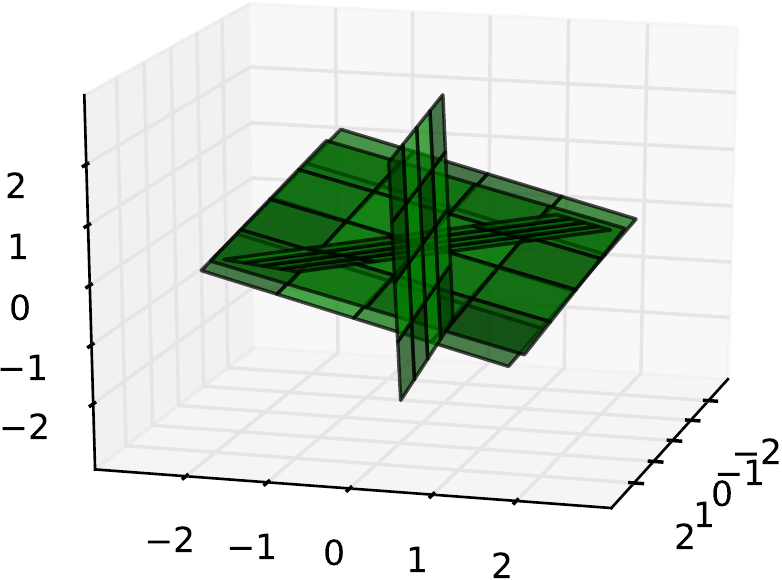}
  }
  \subfigure[intersection]
  {
  \includegraphics[width=0.3\textwidth]{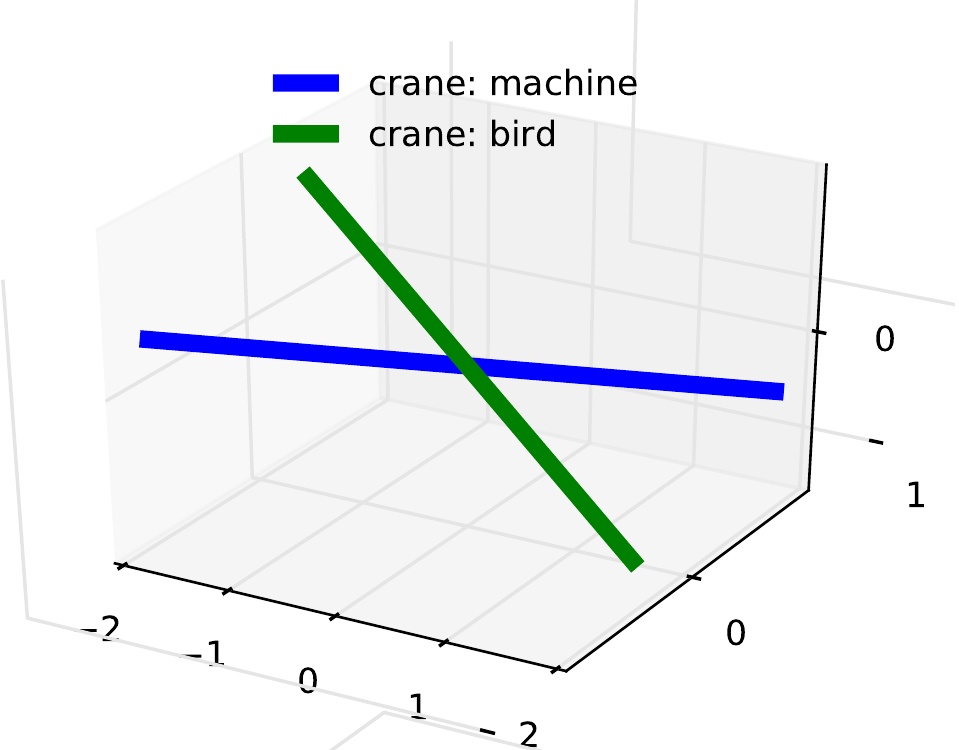}
  }
  \caption{Geometry of contexts for a polysemous word ``crane'': (a) all contexts where ``crane'' means a machine roughly intersect at one direction; (b) all contexts where ``crane'' means a bird roughly intersect at another direction; (c) two directions representing ``crane'' as a machine and as a bird. \label{fig:polysemy}}
\end{figure*}

The property that context subspaces of a monosemous word intersect at one direction  naturally generalizes to polysemy:
\begin{quote}
{\bf Polysemy Intersection Hypothesis}: 
the context subspaces of a polysemous word intersect at different directions for different senses.
\end{quote}
This intuition is validated by the following  experiment, which continues on the same theme as the one done for the monosemous word ``typhoon''. Now   we study the geometry of contexts for a polysemous word ``crane'', which  can either mean a large, tall machine used for moving heavily objects or a tall, long-legged, long-necked bird. We list four contexts for each sense of ``crane'' in Table \ref{tb:polysemy}, repeat the experiment as conducted above for the monosemous word ``typhoon''  and visualize the context subspaces for two senses in Figure \ref{fig:polysemy}(a) and \ref{fig:polysemy}(b) respectively. Figure \ref{fig:polysemy}(c) plots the direction of two intersections. This immediately suggests that the contexts where ``crane'' stands for a bird intersect at one direction and the contexts where ``crane'' stands for a machine, intersect at a different direction as visualized in 3 dimensions. 

\begin{table*}[h]
  \begin{center}
    \begin{tabular}{|p{0.48\textwidth}|p{0.48\textwidth}|}
      \hline
      \multicolumn{1}{|c|}{\bf crane: machine} & \multicolumn{1}{|c|}{\bf crane: bird} \\
      \hline
      In June 1979, the anchor towed ashore and lifted by mobile {\bf crane} into a tank made of concrete built into the ground specifically for the purpose of conserving the anchor.
      & The sandhill crane (``Grus canadensis'') is a species of large {\bf crane} of North America and extreme northeastern siberia. \\
      \hline
      The company ran deck and covered lighters, stick lighters, steam cranes and heavy lift {\bf crane} barges, providing a single agency for Delaware Valley shippers.
      & Although the grey crowned {\bf crane} remains common over much of its range, it faces threats to its habitat due to drainage, overgrazing, and \\
      \hline
      He claimed that his hydraulic {\bf crane} could unload ships faster and more cheaply than conventional cranes. &
      The blue crane (``Anthropoides paradiseus''), also known as the Stanley {\bf crane} and the paradise crane, is the national bird of South Africa.\\
      \hline
      A large pier was built into the harbour to accommodate a heavy lift marine {\bf crane} which would carry the components into the Northumberland Strait to be installed.
      & The sarus {\bf crane} is easily distinguished from other cranes in the region by the overall grey colour and the contrasting red head \\
      \hline
    \end{tabular}
  \end{center}
  \caption{Contexts containing a polysemous word ``crane''\label{tb:polysemy}.}
\end{table*}

\subsection{Sense Induction}
\label{sec:sense-induction}

We can use the representation of senses by the intersection directions of context subspaces for unsupervised sense induction:  supposing the target polysemous word that has $K$ senses (known ahead of time for now), the goal is  to {\em partition} the contexts associated with this target word into $K$ groups  within each of which  the target polysemous word shares the same sense. 
The fact that two groups of context subspaces, corresponding to different senses, intersect at different directions motivates our geometric  algorithm:   we note that each one of the contexts belongs to a group associated by the {\em nearest intersection direction} which serves  as a prototype of the group. Part of the task is also to identify the most appropriate intersection direction vectors associated with each group. This task represents a form of unsupervised clustering which can be formalized as the optimization problem below. 

Given a target polysemous word $w$, $M$ contexts $c_1,...,c_M$ containing $w$ and a number $K$ indicating the number of senses $w$ has, we would like to partition the $M$ contexts into $K$ sets $S_1,...,S_K$ so as to minimize the  distance $d(\cdot,\cdot)$ of each subspace to the intersection direction of its group, 
\begin{align}
L =   \min_{u_1,...,u_K, S_1,...,S_K} \sum_{k=1}^K \sum_{c\in S_k} d^2(u_k, S(c \setminus w)). \label{eq:kmeans}
\end{align}

This problem \eqref{eq:kmeans} is analogous to the objective of $K$-means clustering for vectors and solving it exactly in the worst case can be shown to be NP-hard. We propose a natural algorithm by repurposing traditional $K$-means clustering built for vector spaces to the Grassmannian space as follows (a detailed algorithm chart is provided in Appendix~\ref{app:algo2}):

{\bf Algorithm: $K$-Grassmeans}
\begin{itemize}
\item {\bf Initialization:} we randomly initialize $K$ unit-length vectors $u_1,...,u_K$.
\item {\bf Expectation:} we group contexts based on the distance to each intersection direction: 
 \begin{align*}
    S_k \leftarrow \{c_m:& d(u_k, S(c_m\setminus w)) \le d(u_{k'}, S(c_m\setminus w)) \ \forall k' \}, \forall \ k.
 \end{align*}
\item {\bf Maximization:} we update the intersection direction for each group based on the contexts in the group.
    \begin{align*}
       u_k &\leftarrow \arg \min_{u} \sum_{c\in S_k} d^2(u, S(c\setminus w)), \\
       L &\leftarrow \sum_{k=1}^K \sum_{c\in S_k} d^2(u_k, S(c\setminus w)).
     \end{align*}
\end{itemize}
To ensure good performance, we randomize the intersection directions with multiple different seeds and output the best one in terms of the objective function $L$; this step is analogous to the random initialization conducted in kmeans++ in classical clustering literature \cite{DBLP:journals/jacm/OstrovskyRSS12,DBLP:conf/soda/ArthurV07}.

To get a qualitative feel of this algorithm at work, we consider an exemplar target word ``columbia'' with $K=5$ senses. We considered 100K sentences, extracted from the Wikipedia corpus. The goal of sense induction is  to partition 
the set of contexts into 5 groups, so that within each group the target word ``columbia'' has the same sense. We run $K$-Grassmeans for this target word and extract the intersect vectors $u_1,\ldots u_K$ for $K=5$. One sample sentence for each group is given in Table \ref{tb:induction-example} as an example, from which we can see the first group corresponds to British Columbia in Canada, the second one corresponds to Columbia records, the third one corresponds to Columbia University in New York, the fourth one corresponds to the District of Columbia, and the fifth one corresponds to Columbia River. The performance of $K$-Grassmeans in the context of the target word ``columbia'' is described in detail in the context of sense disambiguation (Section~\ref{sec:disambiguation}). 

\begin{table}[h]
\centering
\begin{tabular}{|c|p{0.8\textwidth}|}
\hline
Group No. &  contexts \\
\hline
1 & (a) research centres in canada it is located on patricia bay and the former british columbia highway a in sidney british {\bf columbia} vancouver island just west of victoria international airport the institute is paired with a canadian coast guard base  \\
\hline
2 & (b) her  big break performing  on the  arthur godfrey show  and had since then  released  a  series  of  successful singles  through {\bf columbia}  records  hechtlancaster  productions  first  published  the   music  from their  film marty  in  april and  june  through  cromwell music this  \\
\hline
3 & (c) fellow at the merrill school of journalism university of maryland  college park  in she was a  visiting scholar  at the {\bf columbia}  university center for the study of human rights in haddad completed a master of international policy  \\
\hline
4 & (d) signed into  law  by  president  benjamin harrison  in  march  a  site  for the  new national conservatory  in the  district  of {\bf columbia} was never  selected  much less  built  the  school continued  to  function  in  new york city   existing solely from philanthropy  \\
\hline
5 & (e) in cowlitz county washington the caples community is located west of woodland along caples road on the east shore of {\bf columbia} river and across the river from columbia city oregon the caples community  is  part  of the  woodland  school district \\
\hline
\end{tabular}
\caption{Semantics of 5 groups for target word ``columbia''.\label{tb:induction-example}}
\end{table}

\begin{figure}[h]
	\centering
    \subfigure[$K$=2]
    {\includegraphics[width=0.3\textwidth]{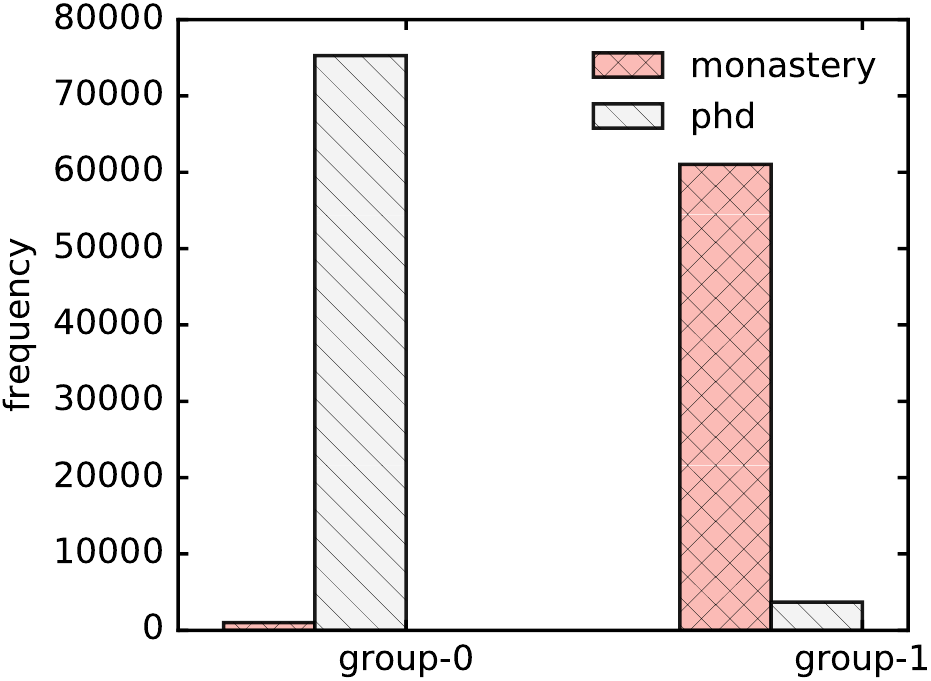}}
    \subfigure[$K$=5]
    {\includegraphics[width=0.3\textwidth]{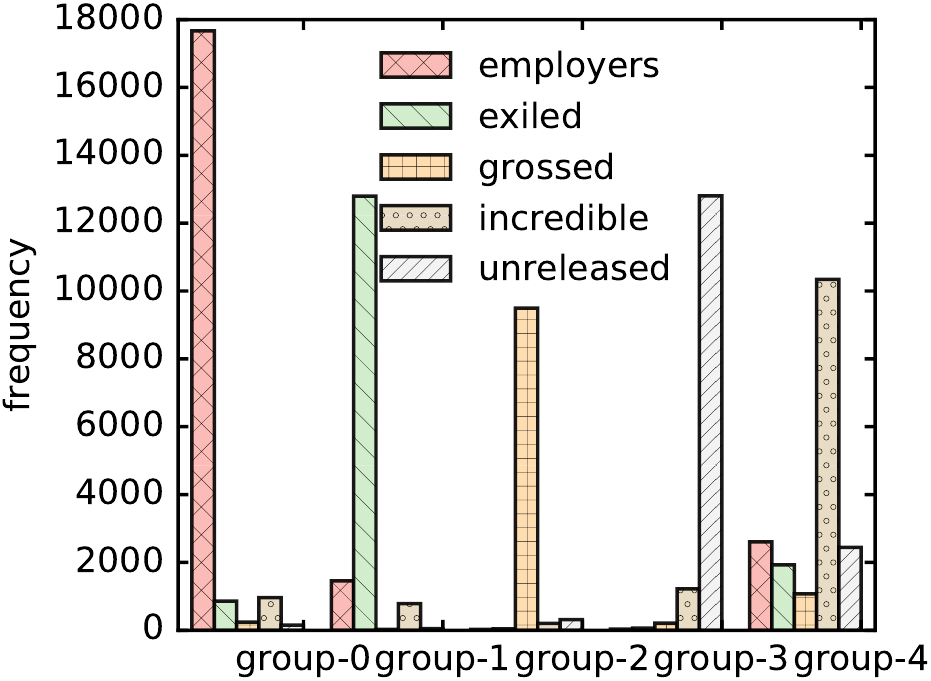}}
    \subfigure[accuracy]
    {\includegraphics[width=0.3\textwidth]{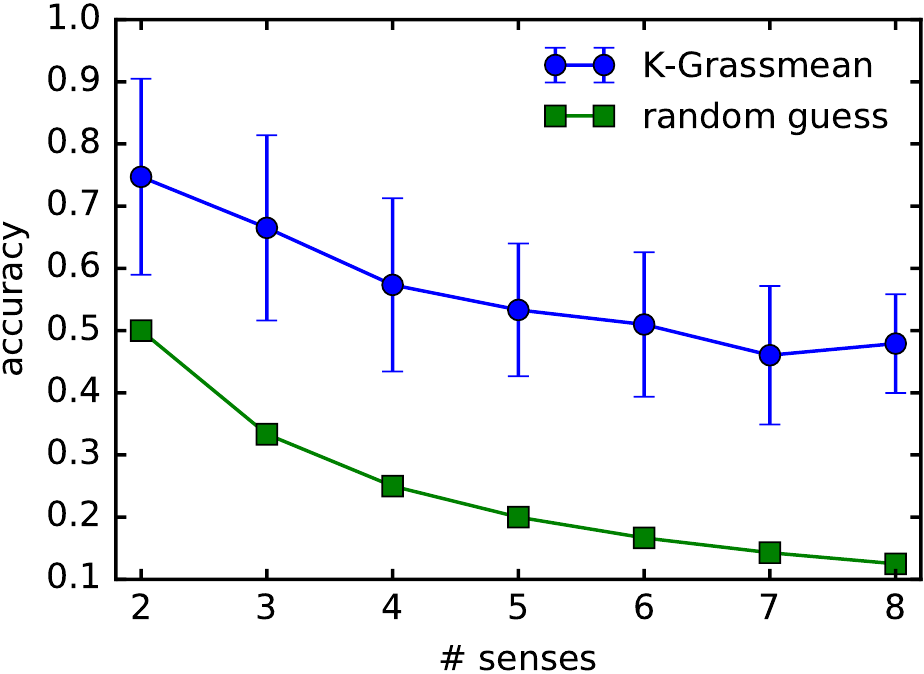}}
    \caption{A synthetic experiment to study the performances of $K$-Grassmeans. \label{fig:artificial-polysemy}: (a) monosemous words: ``monastery'' and ``phd''; (b) $K=5$ monosemous words: ``employers'', ``exiled'', ``grossed'', ``incredible'' and ``unreleased''; (c) accuracy versus $K$.}
\end{figure}

Note that our algorithm can be run for any one {\em specific} target word, and makes for efficient {\em online} sense induction; this is relevant in  information retrieval applications where the sense of the query words may need to be found in real time. To get a feel for how good $K$-Grassmeans is for the sense induction task, we run the following synthetic experiment: we randomly pick $K$ {\em monosemous} words, merge their surface forms to create  a single  {\em artificial polsemous} word, collect all the contexts corresponding to the $K$ monosemous words, replace every occurrence of the $K$ monosemous words by the single artificial polysemous word. Then we run the  $K$-Grassmeans algorithm on these contexts with the artificial polysemous word as the target word, so as to recover their original labels (which are known ahead of time, since we merged known monosemous words together to create the artificial polysemous word).  Figure~\ref{fig:artificial-polysemy}(a)  shows the clustering performances on a realization of the artificial polysemous word made of ``monastery'' and ``phd'' (here $K=2$) and  Figure~\ref{fig:artificial-polysemy}(b))shows the clustering performance when $K=5$ monosemous words  ``employers'', ``exiled'', ``grossed'', ``incredible'' and ``unreleased'' are merged together. We repeat the experiment over 1,00 trials with $K$ varying from 2$\sim$8 and the accuracy of sense induction is reported in Figure~\ref{fig:artificial-polysemy}(c). From these experiments we see that $K$-Grassmeans performs very well, qualitatively and quantitatively. 

A  quantitative experiment on a large and standardized real dataset (which involves real polysemous target words, as opposed to synthetic ones we created), as well as a comparison with other algorithms in the literature, is detailed in Section~\ref{sec:experiments}, where we see that $K$-Grassmeans outperforms state of the art.

\subsection{Sense Disambiguation}
\label{sec:disambiguation}
Having the intersection directions to represent the senses, we are ready to disambiguate a target word sense in a given context using the learned intersection directions specific to this target word: for a new context instance for a polysemous word, the goal is to identify which sense this word means in the context.
Our approach is three-fold:  represent the context by a low dimensional subspace $S(c \setminus w)$ approximation of the linear span of the word embeddings of non-functional words in the context, find the orthogonal projection distance between the intersection vector $u_k(w)$ and the context subspace, and finally output $k^*$ that minimizes the distance, i.e.,
\begin{align}
  k^* = \arg \min_{k} d(u_k(w), S(c \setminus w))\label{eq:disambiguation}.
\end{align}

We refer to \eqref{eq:disambiguation} as a {\em hard} decoding of word senses since this outputs a deterministic label. At times, it makes sense to consider a {\em soft} decoding algorithm  where the output is a probability distribution. The probability that $w$ takes $k$-th sense given the context $c$ is defined via,
\begin{align}
  P(w, c, k) = \frac{\exp(- d(u_k(w), S(c\setminus w)))}{\sum_{k'} \exp(- d(u_{k'}(w), S(c\setminus w)))}. \label{eq:soft-disambiguation}
\end{align}
Here we calculate the probability as a monotonic function of the cosine distance between the intersection vector $u_k(w)$ and the context subspace $S(c\setminus w)$, inspired by similar heuristics in the literature \cite{DBLP:conf/acl/HuangSMN12}.

We applied \eqref{eq:disambiguation} and \eqref{eq:soft-disambiguation} on the target word ``columbia'' and five sentences listed in Table~\ref{tb:induction-example}, the probability distributions $P(w,c,k)$ returned by the soft decoding algorithm and the optimal $k^*$'s returned by hard decoding algorithm are provided in Table~\ref{tb:disambiguation-example}. From Table~\ref{tb:disambiguation-example} we can see that even though the hard decoding algorithm outputs the correct label, some information is missing if we return a single label $k^*$. For example, since we take bag-of-words model in $K$-Grassmean, some words (e.g. ``school'' and ``new york city'' in context (c) provided in Table~\ref{tb:induction-example}) suggest that the meaning for ``columbia'' in this instance might also be Columbia University. The function of those words reflects in the probability distribution returned by the soft decoding algorithm, where we can see the probability that ``columbia'' in this instance means Columbia University is around 0.13. The misleading result mainly comes from the bag-of-words model, and how to resolve it remains open.

\begin{table}[h]
\centering
\begin{tabular}{|c|c|c|c|c|c|c|}
\hline
\multirow{2}{*}{Context No.} & \multirow{2}{*}{hard decoding ($k^*$)} &  \multicolumn{5}{|c|}{soft decoding $P(w, c, k)$} \\
\cline{3-7}
& & $k=1$ & $k=2$ & $k=3$ & $k=4$ & $k=5$ \\
\hline
(a) & 1 & \bf0.81 & 0.01 & 0.01 & 0.05 & 0.13 \\
\hline
(b) & 2 & 0.02 & \bf 0.92 & 0.01 & 0.04 & 0.01  
 \\
\hline
(c) & 3 & 0.01 & 0.00 & \bf 0.91 & 0.06 & 0.01 
 \\
\hline
(d) & 4 & 0.07 & 0.03 & 0.13 & \bf 0.70 & 0.07  \\
\hline
(e) & 5 & 0.04 & 0.01 & 0.01 & 0.05 & \bf 0.90  \\
\hline
\end{tabular}
\caption{Hard decoding and soft decoding for disambiguation of ``columbia'' in five sentences given in Table~\ref{tb:induction-example}.\label{tb:disambiguation-example}}
\end{table}

\section{Lexeme Representation\label{sec:representation}}
  Induction and disambiguation are important tasks by themselves, but several downstream applicatons can use a distributed vector representation of the multiple senses associated with a target word. Just as with word representations, we expect the distributed lexeme representations to have semantic meanings -- similar lexemes should be represented by similar vectors.

  It might seem  natural to represent a lexeme $s_k(w)$ of a given word $w$ by the intersection vector associated with the $k$-th sense group of $w$, i.e., $u_k(w)$. Such an idea is supported by an observation that the intersection vector is close to the word representation vector for many monosemous word. We perform another  experiment to directly confirm this observation: we randomly sample 500 monosemous words which occur at least 10,000 times, for each word we compute the intersection vector and check the cosine similarity between the intersection vector and the corresponding word representation of these monosemous words. We find that on average the  cosine similarity score is  a very high 0.607, with a small standard deviation of 0.095.
  
\begin{figure}[h]
  \centering
  \subfigure[similarity between intersection directions]
  {\includegraphics[width=0.4\textwidth]{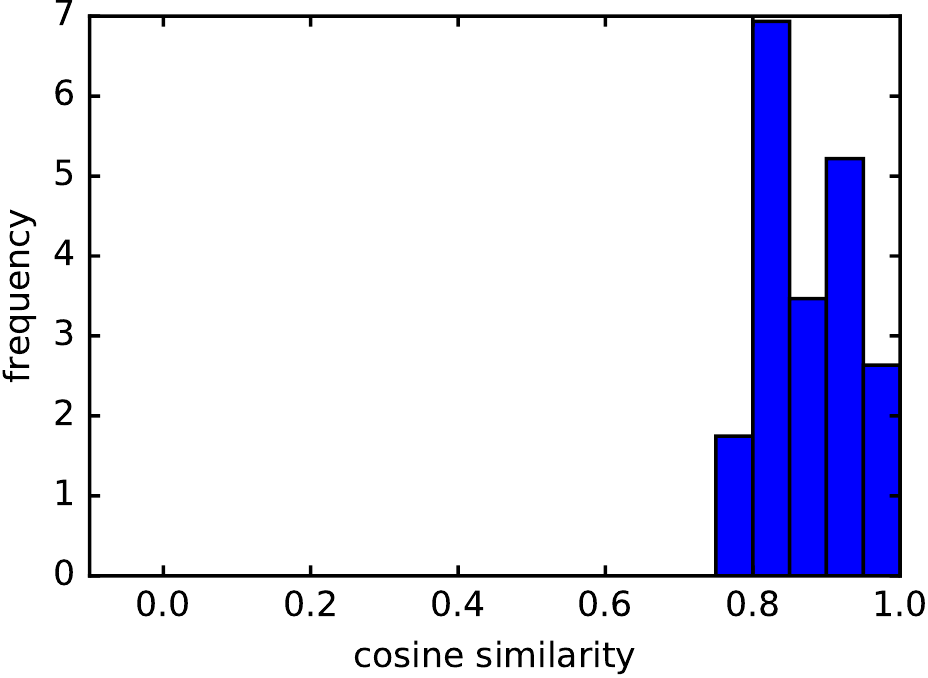}}
  \subfigure[similarity between word representation]
  {\includegraphics[width=0.4\textwidth]{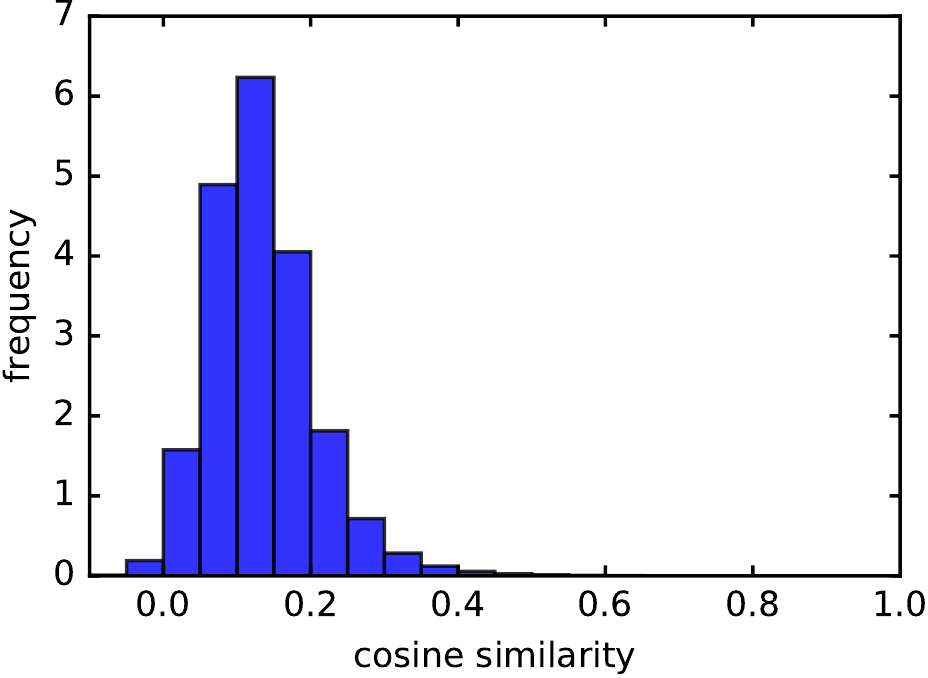}}
  \caption{Cosine similarities between intersection directions and word representations respectively. \label{fig:similarity}}
  \end{figure}

  Despite this empirical evidence, somewhat surprisingly, lexeme representation using the intersection vectors  turns out to be not such a good idea, and the reason is fairly subtle. It turns out that the intersection vectors are concentrated on a relatively small surface area on the sphere (magnitudes are not available in the intersection vectors) --  the cosine similarity between two random intersection vectors among 10,000 intersection vectors (five intersection vectors each for 2,000 polysemous words)
   is 0.889 on average with standard deviation 0.068 (a detailed histogram is provided in Figure \ref{fig:similarity}(a)). This is quite in contrast to analogous statistics for (global) word embeddings from the word2vec algorithm: the cosine similarity between two random word vectors is 0.134 on average with standard deviation 0.072 (a detailed histogram is provided in Figure \ref{fig:similarity}(b)). 
  Indeed, word vector representations are known to be approximately uniformly scattered on the unit sphere (the so-called isotropy property, see  \cite{TACL742}). The intersection vectors cluster together far more and  are quite far from being isotropic -- yet they are still able to distinguish different senses as shown by the empirical studies and qualitative experiments on prototypical examples above (and also on standard datasets, as seen later in Section \ref{sec:experiments}). 
  
  Due to this geometric mismatch between word vectors and intersection directions, and corresponding mismatch in linear algebraic properties expected of these distributed representations,  it is not appropriate to use the intersection direction as the lexeme vector representation.   Why word representations are isotropic but  intersection vectors cluster close to each other is an intriguing open question:  a detailed empirical study of this phenomenon and a theoretical exploration of generative models that might mathematically explain this behavior are exciting future directions of research, and beyond the scope of this paper. In addition to the geometric mismatch, intersection vectors are perhaps not appropriate to represent the lexemes for two further reasons: (a) the intersections only represent directions and lack information about their magnitudes; and (b) the context subspaces are themselves noisy since during the initial phase,   a polysemous word is represented by a single vector. 
 In this light, we propose to learn the lexeme representations by an alternate (and more direct) procedure: first {\em label} the polysemous words in the corpus using the proposed disambiguation algorithm from Section~\ref{sec:disambiguation} and then run a standard word embedding algorithm (we use word2vec) on this labeled corpus, yielding lexeme embeddings. There are several possibilities regarding  labeling and are discussed next.  

\subsection{Hard Decodings} 

We label the corpus using the disambiguation algorithm as in \eqref{eq:disambiguation}. A special label ``IDK'' representing ``I don't know'' is introduced to avoid introducing too many errors during the labeling phase since (a) our approach is based on the bag-of-words model and cannot guarantee to label every sense correctly; (for example, ``arm'' in ``the boat commander was also not allowed to resume his career in the Greek Navy due to his missing {\bf arm} which was deemed a factor that could possibly raise enquiries regarding the mission which caused the trauma.'' will be labeled as ``weapon''); and (b) we are not clear how such errors will affect existing word embedding algorithms.

An ``IDK'' label is introduced via checking the closest distance between the context subspace and the intersection directions, i.e., let $u_{k^*}(w)$ be the closest intersection vector of $w$ to context $c$, we will label this instance as $k^*$ if $d(u_{k^*}(w), S(c\setminus w)) < \theta$ and ``IDK'' otherwise, where $\theta$ is a hyperparameter. A detailed algorithm chart for sense disambiguation and corpus labeling is provided in Appendix~\ref{app:algo3}. 
The ``IDK'' label includes  instances of words that means a rare sense, (for example: ``crane'' as in stretching the neck), or a confusing sense which requires disambiguation of context words (for example: ``book'' and ``ticket'' in ``book a flight ticket''). The IDK labeling procedure is inspired by analogous scenarios in wireless communication  where the log likelihood ratio of (coded) bits is close to zero and in practice are better labeled as ``erasures'', than treating them as informative for the overall decoding task \cite{DBLP:conf/sigcomm/CidonNKV12}.

\subsection{Soft Decodings}

Another way of labeling is via using the absolute scores of $K$-Grassmeans disambiguation for each sense of a target work in a specific context, cf.\ Equation~\eqref{eq:soft-disambiguation}.  Soft decoding involves generating a random corpus by  sampling one sense for every occurrence of a polysemous word according to its probability distribution from \eqref{eq:soft-disambiguation}. Then lexeme representations are obtained via an application of a standard word embedding algorithm (we use word2vec) on this (random) labeled corpus. Since we only consider words that are frequent enough (i.e., whose occurrence is larger than 10,000),  each sense of a polysemous word is sampled enough times to allow a robust lexeme representation with high probability.

Soft decoding benefits in two scenarios: (a) when a context has enough information for disambiguation (i.e., the probability distribution \eqref{eq:soft-disambiguation} concentrates on one), the random sampling will have a high chance making a correct decision. (b) when a context is ambiguous (i.e., the probability distribution have more than one peak), the random sampling will have a chance of not making a wrong (irreversible) decision.

\section{Experiments}
\label{sec:experiments}
Throughout this paper we have conducted multiple qualitative and empirical experiments to highlight and motivate the various geometric representations. In this section we  evaluate our algorithms (on sense  disambiguation method and sense representation)  empirically on (standardized) datasets from the literature, allowing us to get a quantitative feel for the performance on large datasets, as well as afford a comparison with other algorithms from the literature.

\subsection{Preliminaries}
All our algorithms are unsupervised  and operate on a large corpus obtained from Wikipedia dated 09/15. We use WikiExtractor  (\url{http://medialab.di.unipi.it/wiki/Wikipedia_Extractor})  
to extract the plain text. 
We use the skip-gram model from word2vec \cite{DBLP:journals/corr/abs-1301-3781} as the word embedding algorithm where we use the default parameter setting. We set $c=10$ as the context window size and set $N=3$ as the rank of PCA. We choose $K=2$ and $K=5$ in our experiment.
For the disambiguation algorithm, we set $\theta = 0.6$. 

\subsection{Baselines}
Our main comparisons are with algorithms that conduct unsupervised polysemy disambiguation, specifically the sense clustering method of \cite{DBLP:conf/acl/HuangSMN12}, the multi-sense skip gram model (MSSG) of \cite{DBLP:conf/emnlp/NeelakantanSPM14} with different parameters, and the sparse coding method with a global dictionary of \cite{DBLP:journals/corr/AroraLLMR16}. We were able to download the word and sense representations for \cite{DBLP:conf/acl/HuangSMN12,DBLP:conf/emnlp/NeelakantanSPM14} online, and trained the  word and sense representations of \cite{DBLP:journals/corr/AroraLLMR16} on the same corpus as that used by our algorithms. 

\subsection{Sense Induction and Disambiguation}
Word sense induction (WSI) tasks conduct the following test:  given a set of context instances containing a target word, one is asked to partition the context instances into groups such that within each group the target word shares the same sense. We test our induction algorithm, $K$-Grassmeans, on  two  datasets -- one standard and the other custom-built. 
\begin{itemize}
\item {\bf SemEval-2010:} The test set of SemEval-2010 shared task 14 \cite{DBLP:conf/semeval/ManandharKDP10} contains 50 polysemous nouns and 50 polysemous verbs whose senses are extracted from OntoNotes \cite{DBLP:conf/naacl/PradhanX09}, and in total 8,915 instances are expected to be disambiguated. The context instances are extracted from various sources including CNN and ABC. 
\item {\bf Makes-Sense-2016:} Several word senses from SemEval-2010 are too fine-grained in our view (no performance results on tests with native human speakers' is provided in the literature)  -- this creates ``noise'' that that reduces the performance of all the algorithms, and the required senses are perhaps not that useful to downstream applications.  For example, ``guarantee'' (as a noun) is labeled to have four different meanings in the following four sentences:
\begin{itemize}
\item  It has provided a legal {\bf guarantee} to the development of China's Red Cross cause and connections with the International Red Cross movement, signifying that China's Red Cross cause has entered a new historical phase.
\item Some hotels in the hurricane - stricken Caribbean promise money - back {\bf guarantees}.
\item Many agencies roll over their debt , paying off delinquent loans by issuing new loans , or converting defaulted loan {\bf guarantees} into direct loans.
\item Litigation consulting isn't a {\bf guarantee} of a favorable outcome.
\end{itemize}
However, in general they all mean ``a formal promise or pledge''.
 Towards a more human-interpretable version of the WSI task, we  custom-build a dataset whose senses are coarser and clearer. We do this by repurposing a recent dataset created in \cite{DBLP:journals/corr/AroraLLMR16}, as part of their ``police lineup" task. Our dataset contains 50 polysemous words, together with their senses (on average 2.78 senses per word) borrowed from \cite{DBLP:journals/corr/AroraLLMR16}. We generate the testing instances for a target word by extracting all occurrences of it in the Wikipedia Corpus, analyzing its Wikipedia annotations (if  any), grouping those which have the same annotations, and finally merging annotations where the target word shares the same sense. Since the senses are quite readily distinguishable from the perspective of native/fluent  speakers of English, the disambiguation variability, among the human raters we tested our dataset on, is negligible (this effect is also seen in Figure~6 of \cite{DBLP:journals/corr/AroraLLMR16}).
\end{itemize}

We evaluate the performance of the algorithms on this (disambiguation) task  according to standard measures in the literature:  V-Measure \cite{DBLP:conf/emnlp/RosenbergH07} and paired F-Score \cite{DBLP:conf/emnlp/ArtilesAG09}; these two  evaluation metrics also feature in the SemEval-2010 WSI task \cite{DBLP:conf/semeval/ManandharKDP10}.  V-measure is an entropy-based external cluster evaluation metric. Paired F-score evaluates clustering performance by converting the clustering problem into a binary classification problem -- given two instances, do they belong to the same cluster or not?  Both metrics operate on a contingency table $A=\{a_{tk}\}$, where $a_{tk}$ is the number of instances that are manually labeled as $t$ and algorithmically labeled as $k$. A detailed description is given in Appendix~\ref{app:metric} for completeness. Both the metrics  range from 0 to 1, and  perfect clustering gives a score of 1. Empirical statistics show that V-Measure favors those with a larger number of cluster and paired F-score favors those with a smaller number of cluster.

\begin{table}[h]
  \begin{center}
    \begin{tabular}{|r||c|c|c||c|c|c|}
      \hline
      \multirow{2}{*}{\bf algorithms} & \multicolumn{3}{|c||} {\bf SemEval-2010}  & \multicolumn{3}{|c|} {\bf Make-Sense-2016} \\
      \cline{2-7}
      & V-Measure & F-score & \# cluster &  V-Measure & F-score & \# cluster \\
      \hline
     MSSG.300D.30K.key      &         9.00    &       47.26   &       2.88 & 19.40   &       54.49   &       2.88
  \\
      MSSG.300D.6K.key      &           6.90    &       48.43   &       2.45 & 14.40   &       57.91   &       2.35  \\
      \hline
      huang 2012  & 10.60   &       38.05   &       6.63 & 46.86  & 15.9  & 2.74  \\
      \hline
      \#cluster=2  &  7.30    &       \bf 57.14   &       1.93  &  29.30   &       \bf 64.58   &       2.37 \\
      \#cluster=5  & \bf 14.50   &       44.07   &       4.30 & \bf 34.40   &       58.17   &       4.98 \\
      \hline
    \end{tabular}
  \end{center}
  \caption{Performances (V-measure (x100) and paired F-score (x100)) of Word Sense Induction Task on two datasets.}
  \label{tb:wsi}
\end{table}

Table~\ref{tb:wsi} shows the detailed results of our experiments, and from where we  see that $K$-Grassmeans strongly outperforms the others. The main reason behind the better performance seems to be that $K$-Grassmeans disambiguates some subtle senses where the others cannot. For example, following are three sentences containing ``air'':
\begin{itemize}
\item Can you hear me? You're on the {\bf air}.  One of the great moments of live television , isn't it?
\item The {\bf air} force is to take at least 250 more.
\item The empty shells piled here along the roadside fill the {\bf air} with their briny aroma.
\end{itemize}
It can be observed that enough information is contained in the sentence to inform us that the first ``air'' is about broadcasting, the second is about the region above the ground and the third one is about a mixture of gases. $K$-Grassmeans can distinguish all three while the other algorithms cannot. 

\subsection{Lexeme Representation \label{sec:exp:lex-rep}}

The key requirement of lexeme representations should be that they have the same properties as word embeddings, i.e., similar lexemes (or monosemous words) should be represented by similar vectors. Hence we evaluate our lexeme representations on a standard word similarity task focusing on context-specific scenarios:  the Stanford Contextual Word Similarities (SCWS) dataset \cite{DBLP:conf/acl/HuangSMN12}. In addition to word similarity task on SCWS, we also evaluate our lexeme representations on the ``police lineup''
task proposed in \cite{DBLP:journals/corr/AroraLLMR16}.

\paragraph{Word Similarity on SCWS} The task is as follows: given a pair of target words, the algorithm assigns a measure of similarities between this pair of words. The algorithm is evaluated by checking the degree of agreement between the similarity measure given by the algorithm and the similarity measure given by humans in terms of Spearman's rank correlation coefficient. Although this SCWS dataset is not meant specifically for polysemy, we repurpose it for our tasks  since it asks for the similarity between two words in two given sentential contexts (the contexts presumably provide important clues to the human rater on the senses of the target words) and also because this is a large (involving 2,003 word pairs) and standard dataset in the literature  with 10 human rated similarity scores, each rated on an integral scale from 1 to 10. We take the average of 10 human scores to represent the human judgment.

We propose two measures between $w$ and $w'$ given their respective contexts $c$ and $c'$ -- one (denoted by ${\rm HardSim}$) is based on the hard decoding algorithm and the other one (denoted by ${\rm SoftSim}$) is based on the soft one. ${\rm HardSim}$ and ${\rm SoftSim}$ are defined via,
\begin{align*}
  {\rm HardSim}(w, w', c, c') = d(v_k(w), v_{k'}(w')),
\end{align*}
where $k$, and $k'$ are the senses obtained via \eqref{eq:disambiguation}, $v_k(w)$ and $v_{k'}(w)$ are the lexeme representations for the two senses, $d(v, v')$ is the cosine similarity between two vectors $v$ and $v'$, i.e., ($d(v, v') = {v^{\rm T}v}/{\|v\|\|v'\|}$), and
\begin{align*}
  {\rm  SoftSim}(w, w', c, c') = \sum_{k}\sum_{k'}P(w, c, k)P(w', c', k')d(v_k(w), v_{k'}(w')),
\end{align*}
where $P(w, c, k)$ is the probability that $w$ takes $k$-th sense given the context $c$ defined in \eqref{eq:soft-disambiguation}.

Table~\ref{tb:scws} shows the detailed results on this task. 
 Here we conclude that in general our lexeme representations have a similar performance as the state-of-the-art on both soft and hard decisions. It is worth mentioning that the  vanilla  word2vec representation (which simply ignores the provided contexts) also has a comparable performance -- this makes some sense since some of the words in SCWS are monosemous (and hence their meanings are context-dependent). A closer inspection of the results shows that the vanilla word2vec representation  also performs   fairly well  for  polysemous words too -- we hypothesize that this is because the word2vec global representation  is actualy close to the individual sense representations, if the senses occur frequently enough in the corpus.  This hypothesis is supported by the linear algebraic structure uncovered  by  \cite{DBLP:journals/corr/AroraLLMR16}, that a word representation is a linear combination of its lexeme representations: let $v(w) = \sum_{k}\alpha_k v_k(w)$, and let $w'$ be a word semantically close to $k'$-th sense of $w$, then we know that $v(w)^{\rm T}v(w') = \alpha_{k'} v_{k'}(w)^{\rm T} v(w') + \sum_{k\neq k'} \alpha_k v_k(w)^{\rm} v(w')$. Since (a) other senses are irrelevant to $w'$, we can assume $v_k(w)^{\rm} v(w') \approx 0$ for $k\neq k'$, and (b) the $k'$-th sense is frequent enough, we can assume $\alpha_{k'}\approx 1$. Putting (a) and (b) together we can conclude that $v(w)^{\rm T}v(w') \approx v_{k'}(w)^{\rm T}v(w')$ and therefore the inner product between $v(w)$ and $v(w')$ captures the similarity between $k'$-th sense of $w$ and $w'$.

To separate out the effect of the combination of monosemous and polysemous words  in the SCWS dataset, we expurgate monosemous words from the corpus creating a smaller version that we denoted by SCWS-lite. In SCWS-lite, we only consider the pairs of sentences  where {\em both} target words in one pair have the {\em same surface form} but different contexts (and hence potentially different senses). SCWS-lite now contains 241 pairs of words, which is roughly 12\% of the original dataset. 
The performances of the various algorithms are detailed in Table \ref{tb:scws}, from where we see that  our representations (and corresponding algorithms) outperform the state-of-the-art and showcases the superior representational performance when it comes to context-specific  polysemy disambiguation. 

\begin{table}[h]
  \begin{center}
    \begin{tabular}{|r||c|c|c|c|}
      \hline
      \multirow{2}{*}{Model} & \multicolumn{2}{c|}{SCWS} & \multicolumn{2}{c|}{SCWS-lite} \\
      \cline{2-5}
      & SoftSim & HardSim & SoftSim & HardSim \\
      \hline
      skip-gram & \multicolumn{2}{c|}{62.21} & \multicolumn{2}{c|}{--} \\
      \hline
       MSSG.300D.6K & \bf 67.89      &       55.99   & 21.30 &       20.06  \\
      MSSG.300D.30K & 67.84     &       56.66   & 19.49 &       20.67 \\
      NP-MSSG.300D.6K & 67.72   &       58.55   & 20.14 &       19.26\\
      \hline
      Huang 2012 & 65.7 &  26.1 & -- & --\\
      \hline
      \#cluster=2 (hard) & 61.03      &       58.40   & 9.33  &       4.97 \\
      \#cluster=5 (hard) & 60.82      &       53.14   & \bf 27.40 &       \bf 22.28 \\
      \#cluster=2 (soft) & 63.59      &       \bf 63.67   & 5.07  &       6.53 \\
      \#cluster=5 (soft) & 62.46      &       61.23   & 16.54 &       17.83\\
      \hline
    \end{tabular}
  \end{center}
  \caption{Performance (Spearman's rank correlation x100) on SCWS task.\label{tb:scws}.}
\end{table}

\paragraph{Police Lineup Task}
 This task is proposed in \cite{DBLP:journals/corr/AroraLLMR16} to evaluate the efficacy of their sense representations (via the vector representations of the discourse atoms). The testbed contains 200 polysemous words and their 704 senses, where each sense is defined by eight related words. For example, the ``tool/weapon'' sense of ``axes'' is represented by ``handle'', ``harvest'', ``cutting'', ``split'', ``tool'', and ``wood'', ``battle'', ``chop''. The task is the following: given a polysemous word, the algorithm needs to identify the true senses of this word from a group of $m$ senses (which includes many distractors) by outputting $k$ senses from the group. The algorithm is evaluated by the corresponding precision and recall scores.

This task offers another opportunity to test our representations with the others in the literature, and also provide insight into some of those representations themselves. One possible  algorithm is to simply use that proposed in  \cite{DBLP:journals/corr/AroraLLMR16} where we replace their sense representations with ours: Let $s_k(w)$ denote a sense of $w$, and let $L$ denote the set of words representing a sense. We define a similarity score between a sense representation of $w$ and a sense set from the $m$ senses as,
 \begin{align*}
   score(s_k(w), L) = \sqrt{\sum_{w'\in L} (v_{k}(w)^{\rm T} v_w)^2},
 \end{align*}
 and group two senses with highest scores with respect to $s_k(w)$ for each $k$, and then output the top $k$ senses with highest scores. A detailed algorithm is provided in Appendix~\ref{app:algo4}, with a discussion of the potential variants.  

 \begin{figure}[h]
   \centering
   \subfigure[hard]
   {
   \includegraphics[width=0.4\textwidth]{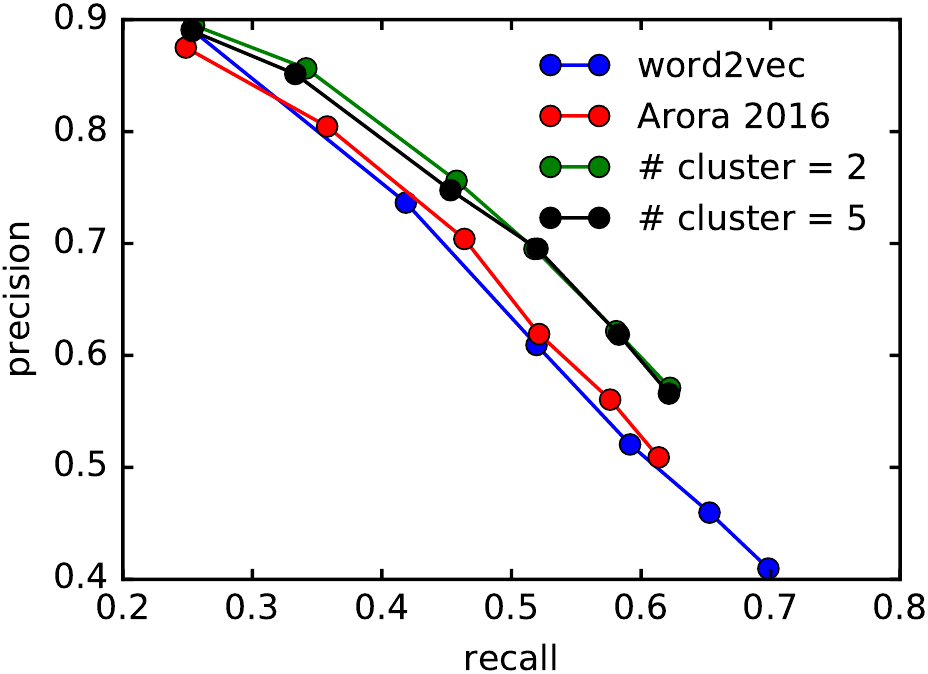}
   }
   \subfigure[soft]
   {\includegraphics[width=0.4\textwidth]{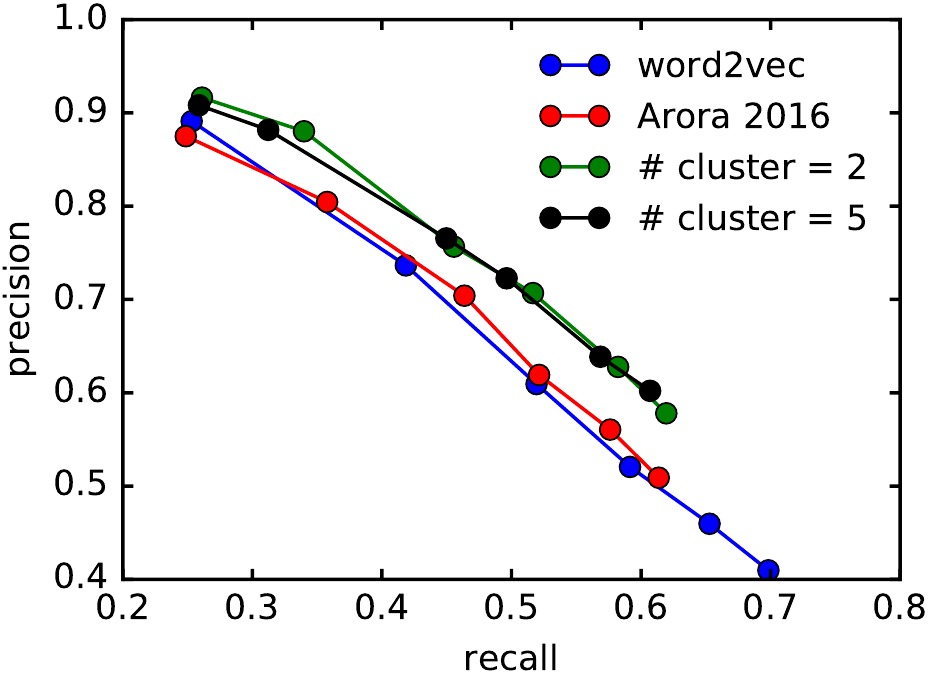}}
   \caption{Precision and recall curve in the sense identification task where we let $m=20$ and $k$ vary from 1 to 6. }\label{fig:princeton}
 \end{figure}

Figure~\ref{fig:princeton} shows the precision and recall curve in the polysemy test where we let $m=20$ and let $k$ vary from 1 to 6. First, we observe that our representations are uniformly better over the precision-recall curve than the state of the art, although by a relatively small margin. Soft decoding performs slightly better than hard decoding over this task. Second, the surprising finding is that  the  baseline we create using vanilla word2vec representations (precise details of this baseline algorithm are  provided in Appendix~\ref{app:algo4} for completeness)  performs as well as the state-of-the-art described in \cite{DBLP:journals/corr/AroraLLMR16}.  A careful look  shows that all algorithm  outputs (word2vec, ours and those in \cite{DBLP:journals/corr/AroraLLMR16}) are highly correlated -- they all make correct calls on obvious instances and all make mistakes for confusing instances. We believe that  this is because of the following: (a) the  algorithm 1 in  \cite{DBLP:journals/corr/AroraLLMR16} is highly correlated with word2vec since their overall similarity measure uses a linear combination of the similarity measures associated with the sense (discourse atom) vector and the word vector (see Step 6 of Algorithm 1 in \cite{DBLP:journals/corr/AroraLLMR16}); (b)  word2vec embeddings appear to have an  ability to  capture  two different senses of a polysemous word (as discussed earlier ); (c) the instances where the errors occur all seem to be either genuinely subtle or rare in the domain where embeddings were trained (for instance ``bat" in the sense of fluttering eyelashes is rare in the Wikipedia corpus, and is one of the error instances).

\section{Related Work \label{sec:relatedwork}}

There are two main approaches to model polysemy: one is supervised and uses linguistic resoures \cite{DBLP:conf/emnlp/ChenLS14,DBLP:conf/acl/RotheS15} and the other is unsupervised  inferring senses directly from a large text corpus  \cite{DBLP:conf/acl/HuangSMN12,DBLP:conf/emnlp/NeelakantanSPM14,DBLP:conf/emnlp/LiJ15,DBLP:journals/corr/AroraLLMR16}. Our approach belongs to the latter category. 

There are  differing approaches to harness hand-crafted lexical resources (such as WordNet):  \cite{DBLP:conf/emnlp/ChenLS14} leverages a ``gloss" as a definition for each lexeme, and uses this to model and disambiguate word senses. \cite{DBLP:conf/acl/RotheS15} models sense and lexeme representations through the {\em ontology} of WordNet. While the approaches are natural and interesting, they are inherently limited due to  (a) the coverage of WordNet: WordNet only covers 26k polysemous words, and the senses for polysemous words are not complete and are domain-agnostic (for example, the mathematical sense for ``ring'' is missing in WordNet and a majority of occurrences of ``ring'' mean  exactly this sense in the Wikipedia corpus) and (b) the fine-grained nature of WordNet: WordNet senses appear at  times too pedantic to be useful in  practical downstream applications (for example, ``air'' in ``This show will {\bf air} Saturdays at 2 P.M.'' and ``air'' in ``We cannot {\bf air} this X-rated song'' are identified to have different meanings).

The unsupervised methods do not suffer from the idiosyncracies of linguistic resources, but are inherently more challenging to pursue since they only rely on the {\em latent} structures of the word senses embedded inside their  contexts. Existing unsupervised approaches can be divided into two categories, based on what aspects of the contexts of target words are used:  (a) global structures of contexts and (b) local structures of contexts.

\begin{figure}[h]
\centering
\includegraphics[width=0.8\textwidth]{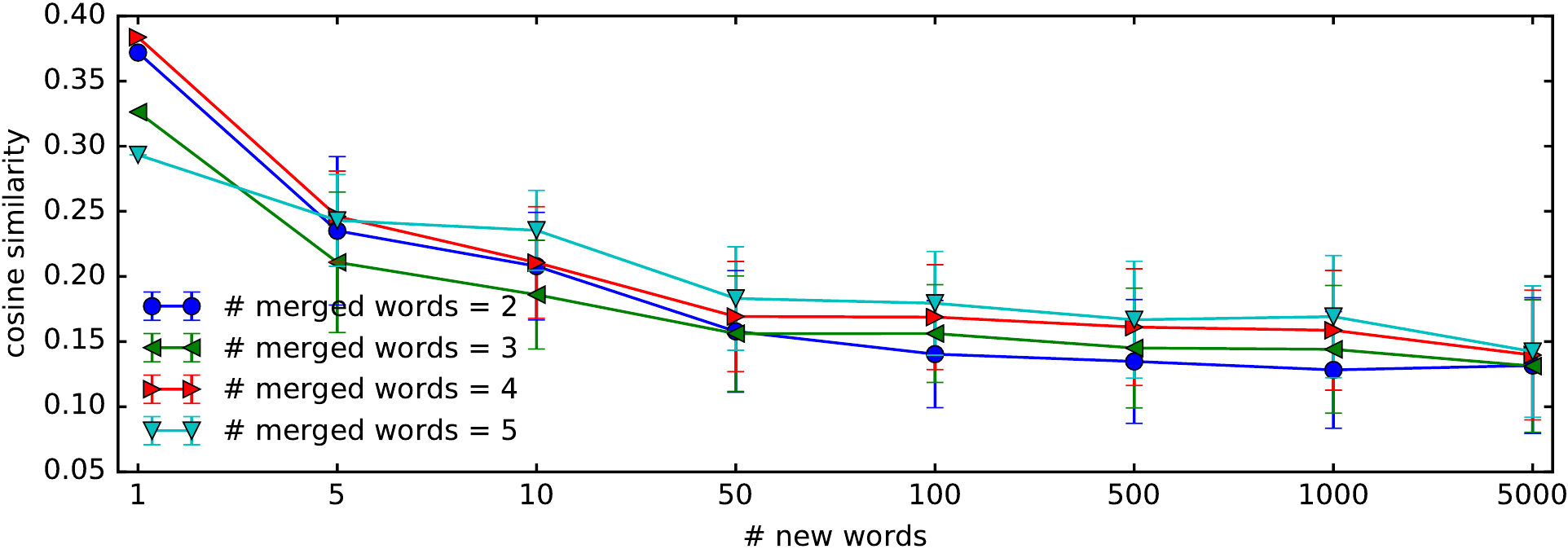}
\caption{A synthetic experiment to study the linear algebraic structure of word senses.\label{fig:linear}}
\end{figure}

{\bf Global structure:} \cite{DBLP:journals/corr/AroraLLMR16} hypothesizes that the global word representation is a linear combination of its sense vectors. This linear algebraic hypothesis  is validated by a surprising  experiment wherein  a single artificial polysemous word is created by merging two random words. The experiment is ingenious and the finding quite surprising but was under a restricted setting: a {\em single} artificial polysemous word is created by merging only {\em two} random words. Upon enlargening these parameters (i.e., many artificial polysemous words are created by merging multiple random words) to better suit the landscape of polysemy in natural language, We find  the linear-algebraic hypothesis to be  fragile: Figure~\ref{fig:linear} plots the linearity fit as a function of the number of artificial polysemous words created, and also as a function of how many words were merged to create any polysemous word.  We see that the linearity fit worsens fairly quickly as the number of polysemous words increases, a scenario that is typical of natural languages. 

The main reason for this effect appears to be that the linearity fit is quite sensitive to the {\em interaction between the word vectors}, caused by the polysemous nature of the words. The linear algebraic hypothesis is mathematically justified  in Section 5 in \cite{DBLP:journals/corr/AroraLLMR16} in terms of the RAND-WALK generative model of \cite{TACL742}  with three extra simplifications. If one  were to generalize this proof to handle  multiple artificial words at the same time, it appears particularly relevant that the simplification 2 should continue to hold. This simplification step involved the assumption that  if $w_1$, ..., $w_n$ be the random words being merged, then (a) $w_i$ and $w_j$ do not occur together in a context window for any $i\neq j$ and (b) any other word $\kappa$  can only occur with a single one of the $w_i$'s in all  context windows. This simplification step clearly no longer holds when $n$ increases, and especially so when $n$ nears  the size of vocabulary. However, this latter scenario (of $n$ being the size of the vocabulary) is the very  basis of of the sparse-coding algorithm proposed in \cite{DBLP:journals/corr/AroraLLMR16} where the latent structure of the multiple senses is modeled as a corpus of discourse atoms where every atom interacts with all the others. 

{The experiment, whose results are depicted in Figure~\ref{fig:linear}, is designed to mimic these   underlying simplifications of the proof in \cite{DBLP:journals/corr/AroraLLMR16}:  we train word vectors via the   skip-gram version of word2vec using the following steps.  (a) We initialize the newly generated artificial polysemous words by random vectors; (b) we initialize, and do not update the (two sets of), vector representations of other words $\kappa$ by the existing word vectors. The embeddings are learnt on  the 2016/06/01 Wikipedia dump, tokenized via   WikiExtractor  (\url{http://medialab.di.unipi.it/wiki/Wikipedia_Extractor});  words that occur less than 1,000 times are ignored; words being merged are chosen randomly in proportion to their frequencies. Due to computational constraints, each instance of mergers is subjected to a single  run of the word2vec algorithm. }

{\bf Local structure:} \cite{DBLP:conf/acl/HuangSMN12,DBLP:conf/emnlp/NeelakantanSPM14} model a context by the {\em average} of its constituent word embeddings and use this average vector as a feature  to induce word senses by partitioning context instances into groups and to disambiguate word senses for new context instances. \cite{DBLP:conf/emnlp/LiJ15} models the senses for a target word in a given context by a Chinese restaurant process,  models the contexts also by averaging its constituent word embeddings and then applies standard word embedding algorithms (continuous bag-of-words (CBOW) or skip-gram). Our approach is broadly similar in spirit to these approaches, in that a local lexical-level model is conceived, but we depart in several ways, the most prominent one being the modeling of the contexts as subspaces (and not as vectors, which is what an average of constituent word embeddings would entail).

\section{Conclusion}

In this paper, we study the geometry of contexts and polysemy and propose a three-fold approach (entitled $K$-Grassmeans) to model target polysemous words in an unsupervised fashion: (a) we represent a context (non-function words surrounding the target word) by a low rank subspace, (b) induce word senses by clustering the subspaces in terms of a distance to an intersection vector and (c) representing lexeme (as a word and sense pair) by labeling the corpus. Our representations are novel and involve nonlinear (Grassmannian) geometry of subspaces and the clustering algorithms are designed to harness this specific geometry.  The overall performance of the method is evaluated quantitatively on standardized word sense induction  and word similarity tasks and we present new state-of-the-art results. Several new avenues of research in natural language representations arise from the ideas in this work and we discuss a few items in detail below. 

\begin{itemize}
\item {\bf Interactions between Polysemous Words}: 
One of the findings of this work, via the experiments in Section~\ref{sec:relatedwork}, is that  polysemous words interact with each other in the corpus. One natural way to harness these intersections, and hence to sharpen $K$-Grassmeans,  is to do an iterative labeling procedure. Both hard decoding and soft decoding (discussed in Section \ref{sec:representation}) can benefit from iterations. In hard decoding, the ``IDK'' labels may be resolved over multiple iterations since (a) the rare senses can become dominant senses once the major senses are already labeled, and (b) a confusing sense can be disambiguated once the polysemous words in its context are disambiguated. In soft decoding, the probability can be expected to concentrate on one sense since each iteration yields yet more  precise context word embeddings. 
This hypothesis is inspired by the success of such procedures inherent in the message passing algorithms for turbo and LDPC codes in reliable wireless communication which share a fair bit of commonality with the setting of polysemy disambiguation \cite{DBLP:books/daglib/0027802}. 
 A quantitative tool to measure the disambiguation improvements from iterations and when to stop the iterative process (akin to the EXIT charts for message passing iterative decoders of LDPC codes \cite{DBLP:journals/tcom/Brink01}) is an interesting research direction,  as is a detailed algorithm design of the iterative decoding and its implementation;  both of these are  beyond the scope of this paper. 

\item {\bf Low Dimensional Context Representation}: A surprising finding of this work is that contexts (sentences) that contain a common target word tend to reside in a low dimensional subspace, as  justified via  empirical observations in Figure~\ref{fig:lowrank}. Understanding this geometrical phenomenon in the context of a generative model (for instance, RAND-WALK of \cite{TACL742} is not able to explain this) is a basic problem of interest, with several relevant applications (including language modeling \cite{DBLP:conf/naacl/IyerOR94,DBLP:journals/aslib/PuciharKMD16} and semantic parsing of sentences (textual entailment, for example \cite{DBLP:conf/mlcw/DaganGM05,bar2006second,giampiccolo2007third})).  Such an understanding could also provide new ideas for the topical subject of representing sentences and paragraphs \cite{DBLP:conf/icml/LeM14,DBLP:journals/corr/WietingBGL15a,DBLP:conf/acl/KenterBR16,DBLP:conf/icml/KusnerSKW15} and eventually combining with document/topic modeling methods such as Latent Dirichlet Allocation \cite{DBLP:journals/jmlr/BleiNJ03}.
\item {\bf Combining Linguistic Resources}: Presently the methods of lexeme representation are either exclusive external resource-based or entirely unsupervised. The unsupervised method of   $K$-Grassmeans reveals robust clusters of senses (and also provides a soft score measuring the robustness (in terms of how frequent the sense is and how sharply/crisply it is used) of  the identified sense). On the other hand, WordNet lists a very detailed number of senses, some  frequent and robust but many others very fine grained; the lack of any accompanying metric that relates to the frequency and robustness of this sense (which could potentially be domain/corpus specific) really makes this resource hard to make computational use of, at least within the context  of polysemy representations. An interesting research direction would be to try to  combine $K$-Grassmeans and existing linguistic resources to automatically define senses of multiple {\em granularities}, along with metrics relating to frequency and robustness of the identified senses.
\end{itemize}

\newpage

\bibliographystyle{plain}
\bibliography{polysemy}

\begin{thebibliography}{10}

\bibitem{TACL742}
Sanjeev Arora, Yuanzhi Li, Yingyu Liang, Tengyu Ma, and Andrej Risteski.
\newblock A latent variable model approach to pmi-based word embeddings.
\newblock {\em Transactions of the Association for Computational Linguistics},
  4:385--399, 2016.

\bibitem{DBLP:journals/corr/AroraLLMR16}
Sanjeev Arora, Yuanzhi Li, Yingyu Liang, Tengyu Ma, and Andrej Risteski.
\newblock Linear algebraic structure of word senses, with applications to
  polysemy.
\newblock {\em CoRR}, abs/1601.03764, 2016.

\bibitem{DBLP:conf/soda/ArthurV07}
David Arthur and Sergei Vassilvitskii.
\newblock k-means++: the advantages of careful seeding.
\newblock In {\em Proceedings of the Eighteenth Annual {ACM-SIAM} Symposium on
  Discrete Algorithms, {SODA} 2007, New Orleans, Louisiana, USA, January 7-9,
  2007}, pages 1027--1035, 2007.

\bibitem{DBLP:conf/emnlp/ArtilesAG09}
Javier Artiles, Enrique Amig{\'{o}}, and Julio Gonzalo.
\newblock The role of named entities in web people search.
\newblock In {\em Proceedings of the 2009 Conference on Empirical Methods in
  Natural Language Processing, {EMNLP} 2009, 6-7 August 2009, Singapore, {A}
  meeting of SIGDAT, a Special Interest Group of the {ACL}}, pages 534--542,
  2009.

\bibitem{bar2006second}
Roy Bar-Haim, Ido Dagan, Bill Dolan, Lisa Ferro, Danilo Giampiccolo, Bernardo
  Magnini, and Idan Szpektor.
\newblock The second pascal recognising textual entailment challenge.
\newblock In {\em Proceedings of the second PASCAL challenges workshop on
  recognising textual entailment}, volume~6, pages 6--4, 2006.

\bibitem{DBLP:journals/jmlr/BengioDVJ03}
Yoshua Bengio, R{\'{e}}jean Ducharme, Pascal Vincent, and Christian Janvin.
\newblock A neural probabilistic language model.
\newblock {\em Journal of Machine Learning Research}, 3:1137--1155, 2003.

\bibitem{DBLP:journals/jmlr/BleiNJ03}
David~M. Blei, Andrew~Y. Ng, and Michael~I. Jordan.
\newblock Latent dirichlet allocation.
\newblock {\em Journal of Machine Learning Research}, 3:993--1022, 2003.

\bibitem{DBLP:conf/emnlp/ChenLS14}
Xinxiong Chen, Zhiyuan Liu, and Maosong Sun.
\newblock A unified model for word sense representation and disambiguation.
\newblock In {\em Proceedings of the 2014 Conference on Empirical Methods in
  Natural Language Processing, {EMNLP} 2014, October 25-29, 2014, Doha, Qatar,
  {A} meeting of SIGDAT, a Special Interest Group of the {ACL}}, pages
  1025--1035, 2014.

\bibitem{DBLP:conf/sigcomm/CidonNKV12}
Asaf Cidon, Kanthi Nagaraj, Sachin Katti, and Pramod Viswanath.
\newblock Flashback: decoupled lightweight wireless control.
\newblock In {\em {ACM} {SIGCOMM} 2012 Conference, {SIGCOMM} '12, Helsinki,
  Finland - August 13 - 17, 2012}, pages 223--234, 2012.

\bibitem{DBLP:conf/mlcw/DaganGM05}
Ido Dagan, Oren Glickman, and Bernardo Magnini.
\newblock The {PASCAL} recognising textual entailment challenge.
\newblock In {\em Machine Learning Challenges, Evaluating Predictive
  Uncertainty, Visual Object Classification and Recognizing Textual Entailment,
  First {PASCAL} Machine Learning Challenges Workshop, {MLCW} 2005,
  Southampton, UK, April 11-13, 2005, Revised Selected Papers}, pages 177--190,
  2005.

\bibitem{Firth1957}
J.~Firth.
\newblock A synopsis of linguistic theory 1930-1955.
\newblock In {\em Studies in Linguistic Analysis}. Philological Society,
  Oxford, 1957.
\newblock reprinted in Palmer, F. (ed. 1968) Selected Papers of J. R. Firth,
  Longman, Harlow.

\bibitem{giampiccolo2007third}
Danilo Giampiccolo, Bernardo Magnini, Ido Dagan, and Bill Dolan.
\newblock The third pascal recognizing textual entailment challenge.
\newblock In {\em Proceedings of the ACL-PASCAL workshop on textual entailment
  and paraphrasing}, pages 1--9. Association for Computational Linguistics,
  2007.

\bibitem{DBLP:conf/acl/HuangSMN12}
Eric~H. Huang, Richard Socher, Christopher~D. Manning, and Andrew~Y. Ng.
\newblock Improving word representations via global context and multiple word
  prototypes.
\newblock In {\em The 50th Annual Meeting of the Association for Computational
  Linguistics, Proceedings of the Conference, July 8-14, 2012, Jeju Island,
  Korea - Volume 1: Long Papers}, pages 873--882, 2012.

\bibitem{DBLP:conf/naacl/IyerOR94}
Rukmini Iyer, Mari Ostendorf, and Jan~Robin Rohlicek.
\newblock Language modeling with sentence-level mixtures.
\newblock In {\em Human Language Technology, Proceedings of a Workshop held at
  Plainsboro, New Jerey, USA, March 8-11, 1994}, 1994.

\bibitem{DBLP:conf/acl/KenterBR16}
Tom Kenter, Alexey Borisov, and Maarten de~Rijke.
\newblock Siamese {CBOW:} optimizing word embeddings for sentence
  representations.
\newblock In {\em Proceedings of the 54th Annual Meeting of the Association for
  Computational Linguistics, {ACL} 2016, August 7-12, 2016, Berlin, Germany,
  Volume 1: Long Papers}, 2016.

\bibitem{DBLP:conf/emnlp/Kim14}
Yoon Kim.
\newblock Convolutional neural networks for sentence classification.
\newblock In {\em Proceedings of the 2014 Conference on Empirical Methods in
  Natural Language Processing, {EMNLP} 2014, October 25-29, 2014, Doha, Qatar,
  {A} meeting of SIGDAT, a Special Interest Group of the {ACL}}, pages
  1746--1751, 2014.

\bibitem{DBLP:conf/icml/KusnerSKW15}
Matt~J. Kusner, Yu~Sun, Nicholas~I. Kolkin, and Kilian~Q. Weinberger.
\newblock From word embeddings to document distances.
\newblock In {\em Proceedings of the 32nd International Conference on Machine
  Learning, {ICML} 2015, Lille, France, 6-11 July 2015}, pages 957--966, 2015.

\bibitem{DBLP:conf/icml/LeM14}
Quoc~V. Le and Tomas Mikolov.
\newblock Distributed representations of sentences and documents.
\newblock In {\em Proceedings of the 31th International Conference on Machine
  Learning, {ICML} 2014, Beijing, China, 21-26 June 2014}, pages 1188--1196,
  2014.

\bibitem{DBLP:conf/nips/LevyG14}
Omer Levy and Yoav Goldberg.
\newblock Neural word embedding as implicit matrix factorization.
\newblock In {\em Advances in Neural Information Processing Systems 27: Annual
  Conference on Neural Information Processing Systems 2014, December 8-13 2014,
  Montreal, Quebec, Canada}, pages 2177--2185, 2014.

\bibitem{DBLP:conf/emnlp/LiJ15}
Jiwei Li and Dan Jurafsky.
\newblock Do multi-sense embeddings improve natural language understanding?
\newblock In {\em Proceedings of the 2015 Conference on Empirical Methods in
  Natural Language Processing, {EMNLP} 2015, Lisbon, Portugal, September 17-21,
  2015}, pages 1722--1732, 2015.

\bibitem{DBLP:conf/semeval/ManandharKDP10}
Suresh Manandhar, Ioannis~P. Klapaftis, Dmitriy Dligach, and Sameer Pradhan.
\newblock Semeval-2010 task 14: Word sense induction {\&}disambiguation.
\newblock In {\em Proceedings of the 5th International Workshop on Semantic
  Evaluation, SemEval@ACL 2010, Uppsala University, Uppsala, Sweden, July
  15-16, 2010}, pages 63--68, 2010.

\bibitem{DBLP:journals/corr/abs-1301-3781}
Tomas Mikolov, Kai Chen, Greg Corrado, and Jeffrey Dean.
\newblock Efficient estimation of word representations in vector space.
\newblock {\em CoRR}, abs/1301.3781, 2013.

\bibitem{DBLP:conf/interspeech/MikolovKBCK10}
Tomas Mikolov, Martin Karafi{\'{a}}t, Luk{\'{a}}s Burget, Jan Cernock{\'{y}},
  and Sanjeev Khudanpur.
\newblock Recurrent neural network based language model.
\newblock In {\em {INTERSPEECH} 2010, 11th Annual Conference of the
  International Speech Communication Association, Makuhari, Chiba, Japan,
  September 26-30, 2010}, pages 1045--1048, 2010.

\bibitem{DBLP:journals/cacm/Miller95}
George~A. Miller.
\newblock Wordnet: {A} lexical database for english.
\newblock {\em Commun. {ACM}}, 38(11):39--41, 1995.

\bibitem{DBLP:conf/icml/MnihH07}
Andriy Mnih and Geoffrey~E. Hinton.
\newblock Three new graphical models for statistical language modelling.
\newblock In {\em Machine Learning, Proceedings of the Twenty-Fourth
  International Conference {(ICML} 2007), Corvallis, Oregon, USA, June 20-24,
  2007}, pages 641--648, 2007.

\bibitem{DBLP:conf/emnlp/NeelakantanSPM14}
Arvind Neelakantan, Jeevan Shankar, Alexandre Passos, and Andrew McCallum.
\newblock Efficient non-parametric estimation of multiple embeddings per word
  in vector space.
\newblock In {\em Proceedings of the 2014 Conference on Empirical Methods in
  Natural Language Processing, {EMNLP} 2014, October 25-29, 2014, Doha, Qatar,
  {A} meeting of SIGDAT, a Special Interest Group of the {ACL}}, pages
  1059--1069, 2014.

\bibitem{DBLP:journals/jacm/OstrovskyRSS12}
Rafail Ostrovsky, Yuval Rabani, Leonard~J. Schulman, and Chaitanya Swamy.
\newblock The effectiveness of lloyd-type methods for the k-means problem.
\newblock {\em J. {ACM}}, 59(6):28, 2012.

\bibitem{DBLP:conf/emnlp/PenningtonSM14}
Jeffrey Pennington, Richard Socher, and Christopher~D. Manning.
\newblock Glove: Global vectors for word representation.
\newblock In {\em Proceedings of the 2014 Conference on Empirical Methods in
  Natural Language Processing, {EMNLP} 2014, October 25-29, 2014, Doha, Qatar,
  {A} meeting of SIGDAT, a Special Interest Group of the {ACL}}, pages
  1532--1543, 2014.

\bibitem{DBLP:conf/naacl/PradhanX09}
Sameer~S. Pradhan and Nianwen Xue.
\newblock Ontonotes: The 90{\%} solution.
\newblock In {\em Human Language Technologies: Conference of the North American
  Chapter of the Association of Computational Linguistics, Proceedings, May 31
  - June 5, 2009, Boulder, Colorado, USA, Tutorial Abstracts}, pages 11--12,
  2009.

\bibitem{DBLP:journals/aslib/PuciharKMD16}
Klen~Copic Pucihar, Matjaz Kljun, John Mariani, and Alan Dix.
\newblock An empirical study of long-term personal project information
  management.
\newblock {\em Aslib J. Inf. Manag.}, 68(4):495--522, 2016.

\bibitem{DBLP:conf/naacl/ReisingerM10}
Joseph Reisinger and Raymond~J. Mooney.
\newblock Multi-prototype vector-space models of word meaning.
\newblock In {\em Human Language Technologies: Conference of the North American
  Chapter of the Association of Computational Linguistics, Proceedings, June
  2-4, 2010, Los Angeles, California, {USA}}, pages 109--117, 2010.

\bibitem{DBLP:books/daglib/0027802}
Thomas~J. Richardson and R{\"{u}}diger~L. Urbanke.
\newblock {\em Modern Coding Theory}.
\newblock Cambridge University Press, 2008.

\bibitem{DBLP:conf/emnlp/RosenbergH07}
Andrew Rosenberg and Julia Hirschberg.
\newblock V-measure: {A} conditional entropy-based external cluster evaluation
  measure.
\newblock In {\em EMNLP-CoNLL 2007, Proceedings of the 2007 Joint Conference on
  Empirical Methods in Natural Language Processing and Computational Natural
  Language Learning, June 28-30, 2007, Prague, Czech Republic}, pages 410--420,
  2007.

\bibitem{DBLP:conf/acl/RotheS15}
Sascha Rothe and Hinrich Sch{\"{u}}tze.
\newblock Autoextend: Extending word embeddings to embeddings for synsets and
  lexemes.
\newblock In {\em Proceedings of the 53rd Annual Meeting of the Association for
  Computational Linguistics and the 7th International Joint Conference on
  Natural Language Processing of the Asian Federation of Natural Language
  Processing, {ACL} 2015, July 26-31, 2015, Beijing, China, Volume 1: Long
  Papers}, pages 1793--1803, 2015.

\bibitem{DBLP:conf/acl/TaiSM15}
Kai~Sheng Tai, Richard Socher, and Christopher~D. Manning.
\newblock Improved semantic representations from tree-structured long
  short-term memory networks.
\newblock In {\em Proceedings of the 53rd Annual Meeting of the Association for
  Computational Linguistics and the 7th International Joint Conference on
  Natural Language Processing of the Asian Federation of Natural Language
  Processing, {ACL} 2015, July 26-31, 2015, Beijing, China, Volume 1: Long
  Papers}, pages 1556--1566, 2015.

\bibitem{DBLP:journals/tcom/Brink01}
Stephan ten Brink.
\newblock Convergence behavior of iteratively decoded parallel concatenated
  codes.
\newblock {\em {IEEE} Trans. Communications}, 49(10):1727--1737, 2001.

\bibitem{DBLP:journals/csl/Veronis04}
Jean V{\'{e}}ronis.
\newblock Hyperlex: lexical cartography for information retrieval.
\newblock {\em Computer Speech {\&} Language}, 18(3):223--252, 2004.

\bibitem{DBLP:journals/corr/WietingBGL15a}
John Wieting, Mohit Bansal, Kevin Gimpel, and Karen Livescu.
\newblock Towards universal paraphrastic sentence embeddings.
\newblock {\em CoRR}, abs/1511.08198, 2015.

\end{thebibliography}

\newpage
\onecolumn
\begin{center}
	\Large Supplementary Material: Geometry of Polysemy
\end{center}
\appendix

\section{Context Representation Algorithm}
\label{app:algo1}

The pseudocode for context representation (c.f. Section \ref{sec:context-representation}) is provided in Algorithm~\ref{algo:representation}.

\begin{algorithm}[h]
\SetKwInOut{Input}{Input}
\SetKwInOut{Output}{Output}
\Input{a context $c$, word embeddings $v(\cdot)$, and a PCA rank $N$. }
Compute the first $N$ principle components of samples $\{v(w'), w'\in c\}$,
\begin{align*}
  u_1,...,u_N &\leftarrow {\rm PCA}(\{v(w'), w'\in c\}), \\
  S &\leftarrow \left\{\sum_{n=1}^N : \alpha_n u_n, \alpha_n \in R\right\}
\end{align*} \\
\Output{$N$ orthonormal basis $u_1, ..., u_N$ and a subspace $S$.},
\caption{The algorithm for context representation.}
\label{algo:representation}
\end{algorithm}

\section{Sense Induction Algorithm}
\label{app:algo2}

The pseudocode for word sense induction (c.f. Section \ref{sec:sense-induction}) is provided in Algorithm~\ref{algo:induction}.

\begin{algorithm}[h]
\SetKwInOut{Input}{Input}
\SetKwInOut{Output}{Output}
\Input{contexts $c_1,...,c_M$, and integer $K$. }
Initialize unit length vectors $u_1,...,u_K$ randomly, initialize $L\leftarrow \infty$, \\
\While{$L$ does not converge}
{
 {\bf Expectation step}: assign subspaces to the group which has the closest intersection direction,
 \begin{align*}
    S_k \leftarrow \{c_m: d(u_k, S(c_m\setminus w)) \le d(u_{k'}, S(c_m\setminus w)) \ \forall k' \},
 \end{align*} \\
 {\bf Maximization Step}: update the new intersection direction that minimize the distances to all subspace in this group as \eqref{eq:mono-opt}.
 \begin{align*}
   u_k &\leftarrow \arg \min_{v} \sum_{c\in S_k} d^2(u, S(c\setminus w)), \\
   L &\leftarrow \sum_{k=1}^K \sum_{c\in S_k} d^2(u_k, S(c\setminus w)),
 \end{align*}
}
\Output{$K$ intersection directions $v_1, ..., v_K$.}
\caption{Word sense induction algorithm with given number of senses.}\label{algo:induction}
\end{algorithm}

\section{Sense Disambiguation Algorithm}
\label{app:algo3}

The pseudocode for the word sense disambiguation (c.f. Section~\ref{sec:disambiguation}) is provided in Algorithm~\ref{algo:disambiguation}.
\begin{algorithm}[h]
\SetKwInOut{Input}{Input}
\SetKwInOut{Output}{Output}
\Input{a context $c$, word embeddings $v(\cdot)$, a PCA rank $N$, a set of intersection directions $u_k(w)$'s, and a threshold $\theta$. }
Compute denoised context subspace, $$S\leftarrow S(c\setminus w),$$ \\
Compute the distance between $S(c\setminus w)$ and intersections, 
$$d_k \leftarrow d(u_k, S),$$ \\
\If{hard decoding}{
Get the closest cluster,
$$k^* \leftarrow \arg \min d_k,$$ \\
Check the threshold,\\
\If{$d_{k^*} < \theta$}
{
    \Return{$k^*$};
}
\Return{IDK};}
\If{soft decoding}{
	Compute the probabilities:
    	$$P(w, c, k) \leftarrow \frac{\exp(-10 d(u_k(w), S(c\setminus w)))}{\sum_{k'} \exp(-10 d(u_{k'}(w), S(c\setminus w)))}, $$
}
\Return{$P(w,c,k)$ for $k=1,...,K$;} \\
\Output{An label indicating which sense this word means in this context.}
\caption{The algorithm for sense disambiguation.}\label{algo:disambiguation}
\end{algorithm}

\section{V-Measure and Paired F-score}
\label{app:metric}

Clustering algorithms partition $N$ data points into a set of clusters $\mathcal{K}=\{1,2,...,K\}$. Given the ground truth, i.e., another partition of data into another set of clusters $\mathcal{T} = \{1,2,...,T\}$, the performance of a clustering algorithm is evaluated based on a contingency table $A = a_{tk}$ representing the clustering algorithm, where $a_{tk}$ is the number of data whose ground truth label is $t$ and algorithm label is $k$. There are two intrinsic properties of all desirable evaluation measures:  
\begin{itemize}
	\item The measure should be permutation-invariant, i.e., the measure should be the same if we permutate the labels in $\mathcal{K}$ or $\mathcal{T}$.
    \item The measure should encourage   intra-cluster similarity and penalize  inter-cluster similarity.
\end{itemize}

V-Measure \cite{DBLP:conf/emnlp/RosenbergH07} and paired F-Score \cite{DBLP:conf/emnlp/ArtilesAG09} are two standard measures, the definitions of which are given below.

\subsection{V-Measure}

V-Measure is an entropy-based metric, defined as a harmonic mean of homogeneity and completeness. 
\begin{itemize}
\item Homogeneity is satisfied if the data points belong to one algorithm cluster fall into a single ground truth cluster, formally defined as,
\begin{align*}
	h = \left\{
    \begin{array}{ll} 
    	1 & \textrm{ if $H(\mathcal{T}) =0 $ } \\ 
        1 - \frac{H(\mathcal{T}|\mathcal{K})}{H(\mathcal{T})} & \textrm{ otherwise,}
    \end{array} 
    \right.
\end{align*}
where 
\begin{align*}
	H(\mathcal{T} | \mathcal{K}) &= - \sum_{k=1}^K \sum_{t=1}^ T a_{tk}{N} \log \frac{a_{tk}}{\sum_{t=1}^T} a_{tk}, \\
    H(\mathcal{T}) &= -\sum_{t=1}^T \frac{\sum_{k=1}^K a_{tk}}{N} \log \frac{\sum_{k=1}^k a_{ck}}{N}.
\end{align*}
\item Completeness is analogous to homogeity. Formally this is defined as: 
\begin{align*}
	c = \left\{
    \begin{array}{ll} 
    	1 & \textrm{ if $H(\mathcal{K}) =0 $ } \\ 
        1 - \frac{H(\mathcal{K}|\mathcal{T})}{H(\mathcal{K})} & \textrm{ otherwise,}
    \end{array} 
    \right.
\end{align*}
where
\begin{align*}
	H( \mathcal{K} | \mathcal{T}) &= -  \sum_{t=1}^T \sum_{k=1}^K a_{tk}{N} \log \frac{a_{tk}}{\sum_{k=1}^K} a_{tk}, \\
    H(\mathcal{K}) &= -\sum_{k=1}^K \frac{\sum_{t=1}^T a_{tk}}{N} \log \frac{\sum_{t=1}^T a_{ck}}{N}.
\end{align*}
\end{itemize}
Given $h$ and $c$, the V-Measure is their harmonic mean, i.e.,  
\begin{align*}
	V = \frac{2hc}{h+c}.
\end{align*}

\subsection{Paired F-score}

Paired F-score evaluates clustering performance by converting the clustering into a binary classification problem -- given two instances, do they belong to the same cluster or not? 

For each cluster $k$ identified by the algorithm, we generate ${{ \sum_{t=1}^T a_{tk}} \choose 2}$ instance pairs, and for each ground true cluster, we generate ${{ \sum_{k=1}^K a_{tk}} \choose 2}$ instances pairs. Let $F(\mathcal{K})$ be the set of instance pairs from the algorithm clusters and let $F(\mathcal{T})$ be the set of instances pairs from ground truth clusters. Precision and recall is defined accordingly: 
\begin{align*}
	P = \frac{|F(\mathcal{K})\cap F(\mathcal{T})|}{F(\mathcal{K})}, \\
    R = \frac{|F(\mathcal{K})\cap F(\mathcal{T})|}{F(\mathcal{S})},
\end{align*}
where $|F(\mathcal{K})|$, $|F(\mathcal{T})|$ and $|F(\mathcal{K})\cap F(\mathcal{T})|$  can be computed using the matrix $A$ as below: 
\begin{align*}
	|F(\mathcal{K})| &= \sum_{k=1}^K {{ \sum_{t=1}^T a_{tk}} \choose 2}, \\
    |F(\mathcal{T})| &= \sum_{t=1}^T {{ \sum_{k=1}^K a_{tk}} \choose 2}, \\
    |F(\mathcal{K})\cap F(\mathcal{T})| &= \sum_{t=1}^T \sum_{k=1}^K {{a_{tk}} \choose 2}.
\end{align*}

\section{Algorithms for Police Lineup Task}
\label{app:algo4}

We first introduce the baseline algorithm for word2vec and our algorithm, as given in Algorithm~\ref{algo:word2vec} and \ref{algo:our}. Both algorithms are motivated by the algorithm in \cite{DBLP:journals/corr/AroraLLMR16}.

In our algorithm, the similarity score can be thought as a mean value of the word similarities between a target word $w$ and a definition word $w'$ in the given sense $L$, we take the power mean with $p=2$. Our algorithm can be adapted to different choice of $p$, i.e.,
\begin{align*}
  score_p(s_k(w), L) \leftarrow \left(\sum_{w'\in L} \left|v_{k}(w)^{\rm T} v_w\right|^p\right)^{1/p}  - \frac{1}{|V|} \sum_{w'' \in V} \left(\sum_{w'\in L} \left|v_{k}(w'')^{\rm T} v_w\right|^p\right)^{1/p} 
\end{align*}
Different choice of $p$ leads to different preferences of the similarities between $w$ and $w'\in L$, generally speaking larger weights put on relevant words with larger $p$:
\begin{itemize}
	\item If we take $p=0$, $score_1(w, L)$ turns to be an average of the similarities;
    \item If we take $p = \infty$, $score_{\infty}(w, L)$ measures the similarity between $w$ and $L$ by the similarity between $w$ and the most relevant word $w'\in L$, i.e.,
    \begin{align*}
    	score_{\infty}(s_k(w), L) \leftarrow \max_{w'\in L} |v_k(w)^{\rm T} v(w')|
    \end{align*}
\end{itemize}
In our case we take $p=2$ to allow enough (but not too much) influence from the most relevant words.

\begin{algorithm}[htbp]
\SetKwInOut{Input}{Input}
\SetKwInOut{Output}{Output}
\Input{a target word $w$, $m$ senses $L_1,...,L_m$, word vectors $v(w')$ for every $w'$ in the vocabulary $V$.}
Compute a similarity score between $w$ and a sense $L_i$,
$$score(w, L_i) \leftarrow \sqrt{\sum_{w' \in L_i} \left(v(w)^{\rm T} v(w')\right)^2} - \frac{1}{|V|}\sum_{w''} \sqrt{\sum_{w' \in L_i} \left(v(w'')^{\rm T} v(w')\right)^2} $$ \\
\Return{Top $k$ $L$'s with highest similarity scores.} \\
\Output{$k$ candidate senses.}
\caption{The baseline algorithm with word2vec for Princeton Police Lineup Task.}\label{algo:word2vec}
\end{algorithm}

\begin{algorithm}[htbp]
\SetKwInOut{Input}{Input}
\SetKwInOut{Output}{Output}
\Input{a target word $w$, $m$ senses $L_1,...,L_m$, lexeme representations $v_k(w)$ for every sense of $w$, and word representations $v(w')$ for $w'$ in the vocabulary.}
candidates $\leftarrow$ list() \\
scores $\leftarrow$ list() \\
\For{every sense $s_k(w)$ of $w$}
{
	\For{every $i=1,2,...,m$}
    {
    Compute a similarity score between a lexeme $s_k(w)$ and a sense $L_i$,
    \begin{align*}
    	score(s_k(w), L) \leftarrow & \sqrt{\sum_{w'\in L} (v_{k}(w)^{\rm T} v_{w'})^2}   \\
        & +  \sqrt{\sum_{w'\in L} (v(w)^{\rm T} v(w'))^2} - \frac{1}{|V|}\sum_{w''\in V} \sqrt{\sum_{w'\in L} (v(w'')^{\rm T} v(w'))^2} , 
    \end{align*}
    }
    $L_{i_1}, L_{i_2} \leftarrow$ top 2 senses.  \\
    candidates $\leftarrow$ candidates + $L_{i_1}$ +  $L_{i_2}$ \\
    scores $\leftarrow$ scores + $score(s_k(w), L_{i_1})$ + $score(s_k(w), L_{i_1})$ \\ 
}
\Return{Top $k$ $L$'s with highest similarity scores.} \\
\Output{$k$ candidate senses.}
\caption{Our algorithm for Princeton Police Lineup Task.}\label{algo:our}
\end{algorithm}

Both our algorithm and the algorithm in \cite{DBLP:journals/corr/AroraLLMR16} do not take into account that one atom represents one sense of the target word, and thus some atoms might generate two senses in the output $k$ candidates. A more sophisticated algorithm is required to address this issue.

\end{document}